\documentclass[runningheads]{llncs}
\usepackage{graphicx}
\usepackage{comment} 
\usepackage{amsmath,amssymb} %
\usepackage{color}
\usepackage{amsmath}
\usepackage{wrapfig}
\usepackage[nospace]{cite}
\usepackage[width=122mm,left=12mm,paperwidth=146mm,height=193mm,top=12mm,paperheight=217mm]{geometry}
\usepackage[hidelinks]{hyperref}

\usepackage{xcolor}

\newcommand{\transform}[2]{T_{#1 #2}}
\newcommand{\SE}{\mathrm{SE}(3)}

\newcommand{\argmin}{\operatorname*{argmin}}

\newcommand{\symdist}[1]{D_{#1}}

\usepackage{subcaption}
\begin{document}
\pagestyle{headings}
\mainmatter
\def\ECCVSubNumber{2838}

\title{CosyPose: Consistent multi-view multi-object 6D pose estimation}

\titlerunning{CosyPose: Consistent multi-view multi-object 6D pose estimation}
\author{Yann Labb\'e\inst{1,2} \and
  Justin Carpentier\inst{1,2} \and
  Mathieu Aubry\inst{3} \and
  Josef Sivic\inst{1,2,4}}
\authorrunning{Y. Labb\'e et al.}

\institute{
  \'Ecole normale sup\'erieure, CNRS, PSL Research University, Paris, France \and 
  INRIA, Paris \and
  LIGM, \'Ecole des Ponts, Univ Gustave Eiffel, CNRS, Marne-la-vall\'ee, France \and
  Czech Institute of Informatics, Robotics and Cybernetics, Czech Technical University in Prague\\
}

\maketitle

\begin{abstract}
We introduce an approach for recovering the 6D pose of multiple known objects in a scene captured by a set of input images with unknown camera viewpoints.  %
First, we present a single-view single-object 6D pose estimation method, which we use to generate 6D object pose hypotheses. Second, we develop a robust method for matching individual 6D object pose hypotheses across different input images in order to jointly estimate camera viewpoints and 6D poses of all objects in a {\em single consistent scene}. Our approach explicitly handles object symmetries, does not require depth measurements, is robust to missing or incorrect object {hypotheses}, and automatically recovers the number of objects in the scene.
Third, we develop a method for global scene refinement given multiple object hypotheses and their correspondences across views. This is achieved by solving an {\em object-level bundle adjustment} problem that refines the poses of cameras and objects to minimize the reprojection error in all views. 
We demonstrate that the proposed method, dubbed CosyPose, outperforms current state-of-the-art results for single-view and multi-view 6D object pose estimation by a large margin on two challenging benchmarks: the YCB-Video and T-LESS datasets. Code and pre-trained models are available \href{https://www.di.ens.fr/willow/research/cosypose/}{on the project webpage.\footnote{\href{https://www.di.ens.fr/willow/research/cosypose/}{https://www.di.ens.fr/willow/research/cosypose/}} }

\end{abstract}

\section{Introduction}

\begin{figure}[t]
  \centering

   \newcommand\teaserheight{3.5cm}
   \begin{subfigure}[t]{0.442\textwidth}
        \centering
        \includegraphics[width=\textwidth]{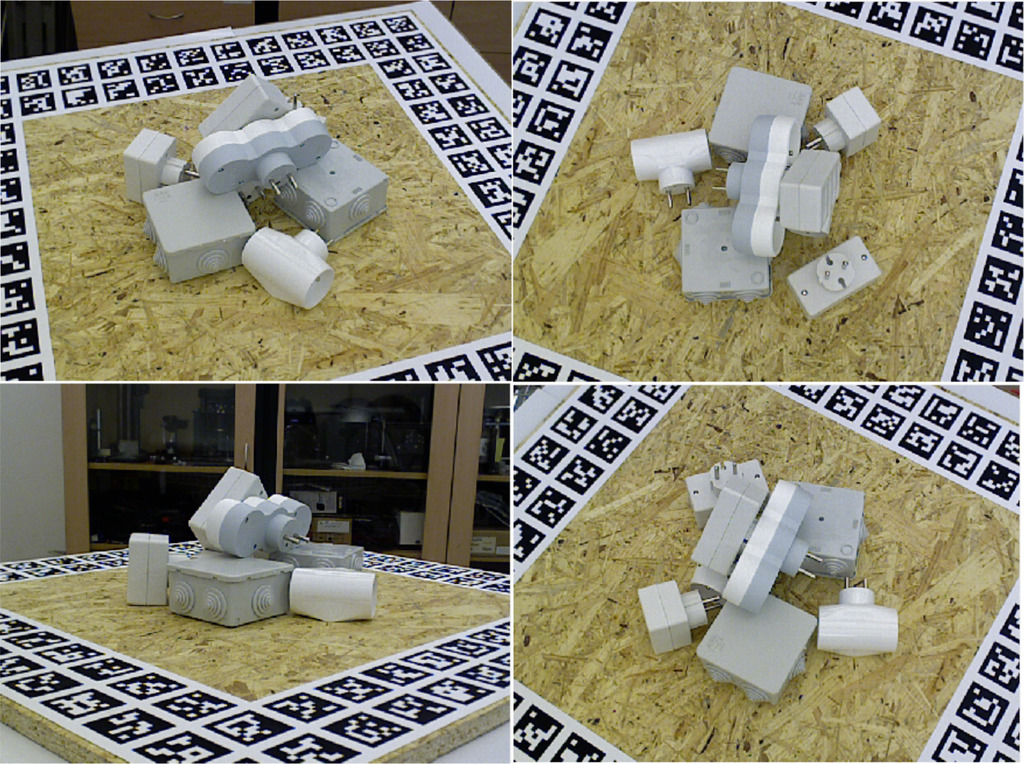}
        \caption{Input: RGB images.}
    \end{subfigure}%
   \begin{subfigure}[t]{0.56\textwidth}
        \centering
        \includegraphics[width=\textwidth]{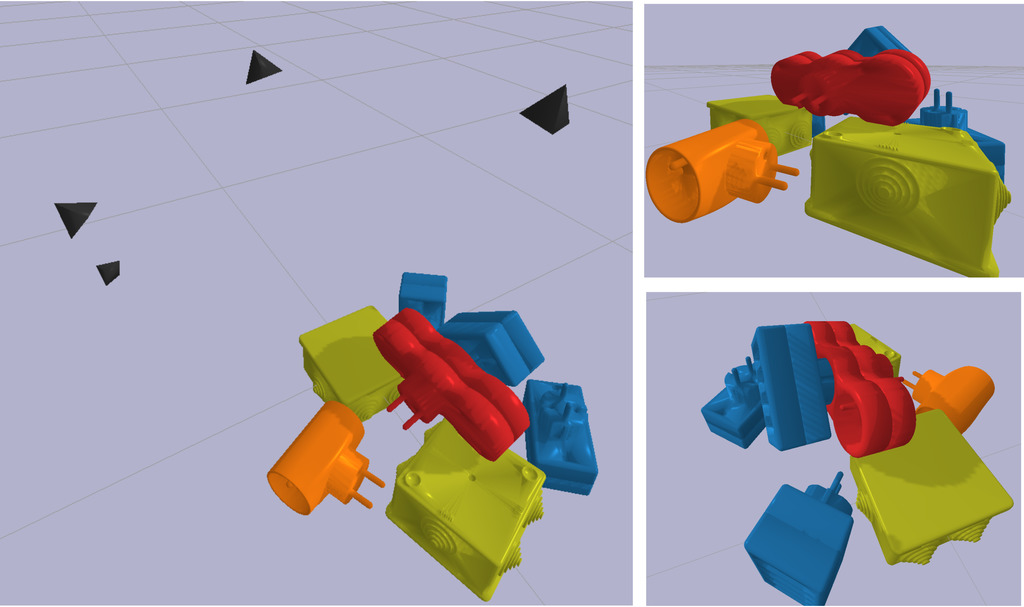}
        \caption{Output: full scene model including objects and camera poses.}
    \end{subfigure}%

  \caption{{\bf CosyPose: 6D object pose estimation optimizing multi-view COnSistencY.} Given (a) a set of RGB images depicting a
  scene with known objects taken from unknown viewpoints,
  our method accurately reconstructs the scene, (b) recovering all objects in the scene, their 6D pose and the camera viewpoints. Objects are  enlarged for the purpose of visualization.}
  \label{fig:teaser}
\end{figure}

The goal of this work is to estimate accurate 6D poses of multiple known objects in a 3D scene captured by multiple cameras with unknown positions, as illustrated in Fig.~\ref{fig:teaser}. 
This is a challenging problem because of the texture-less nature of many objects, the presence of multiple similar objects, the unknown number and type of objects in the scene, and the unknown positions of cameras. 
Solving this problem would have, however, important applications in robotics where the knowledge of accurate position and orientation of objects within the scene would allow the robot to plan, navigate and interact with the environment. %

Object pose estimation is one of the oldest computer vision problems \cite{Roberts1963-ck,Lowe1987-yf,Lowe1999-bf}, yet it remains an active area of research~\cite{Rad2017-de,Peng2018-ev,Tremblay2018-bd,Park2019-od,Zakharov2019-fx,Wang2019-ml,Li2018-fp,Wang2019-bb}.
The best performing methods that operate on RGB (no depth) images 
\cite{Li2018-fp,Park2019-od,Zakharov2019-fx,Wang2019-bb,Sundermeyer2018-es}
are based on trainable convolutional neural networks and are able to deal with symmetric or textureless objects, which were challenging for earlier methods relying on local~\cite{Lowe1999-bf,Bay2006-hj,Hinterstoisser2011-es,Collet2010-zj,Collet2011-lj}
or global \cite{Dalal2005-nd} gradient-based image features. %
However, most of these works consider objects independently and estimate their poses using a single input (RGB) image. 
Yet, in practice, scenes are composed of many objects and multiple images of the scene are often available, e.g. obtained by a single moving camera, or in a multi-camera set-up. 
In this work, we address these limitations
and develop an approach that combines information from {\em multiple views} and estimates jointly the pose of {\em multiple objects} to obtain a single consistent scene interpretation.

While the idea of jointly estimating poses of multiple objects from multiple views may seem simple, the following challenges need to be addressed.   
First, object pose hypotheses made in individual images cannot easily be expressed in a common reference frame when the relative transformations between the cameras are unknown. This is often the case in practical scenarios where camera calibration cannot easily be recovered using local feature registration because the scene lacks texture or the baselines are large.
Second, the single-view 6D object pose hypotheses have gross errors in the form of false positive and missed detections. 
Third, the candidate 6D object poses estimated from input images
are noisy as they suffer from depth ambiguities inherent to single view methods.

In this work, we describe an approach that addresses these challenges. We start from 6D object pose hypotheses that we estimate from each view using a new render-and-compare approach inspired by DeepIM~\cite{Li2018-fp}. 
First, we match individual object pose hypotheses across different views and use the resulting {\em object-level} correspondences to recover the relative positions between the cameras. Second, gross errors in object detection are addressed using a robust object-level matching procedure based on RANSAC, optimizing the overall scene consistency. Third, noisy single-view object poses are significantly improved using a global {\em refinement procedure} based on object-level bundle adjustment. The outcome of our approach that optimizes multi-view COnSistencY, hence dubbed CosyPose, is a single consistent reconstruction of the input scene. {
Our single-view single-object pose estimation method obtains state-of-the-art results on the YCB-Video \cite{Xiang2018-dv} and T-LESS \cite{Hodan2017-pq} datasets, achieving a significant 34.2\% absolute improvement over the state-of-the-art \cite{Park2019-od} on T-LESS. Our multi-view framework clearly outperforms \cite{Li2018-vl} on YCB-Video while not requiring known camera poses and not being limited to a single object of each class per scene. On both datasets, we show that our multi-view solution significantly improves pose estimation and 6D detection accuracy over our single-view baseline.}

\section{Related work}
\label{sec:related_work}

Our work builds on results in single-view and multi-view object 6D pose estimation
from RGB images and object-level SLAM.%

\paragraph{Single-view single-object 6D pose estimation.} 
The object pose estimation problem~\cite{Collet2010-zj,Collet2011-lj}  has been approached either by estimating the pose from 2D-3D correspondences using local invariant features~\cite{Lowe1999-bf,Bay2006-hj}, or directly by estimating the object pose using  template-matching~\cite{Hinterstoisser2011-es}. However, local features do not work well for texture-less objects and global templates often fail to detect partially occluded objects.   
Both of these approaches (feature-based and template
matching) have been revisited using deep neural networks. 
A convolutional neural network (CNN) can be used to detect object
features in 2D~\cite{Rad2017-de,Kehl2017-ek,Tremblay2018-bd,Tekin2017-hp,Xiang2018-dv} or to
directly find 2D-to-3D correspondences~\cite{Park2019-od,Zakharov2019-fx,Peng2018-ev,Pitteri2019-sn}.
Deep approaches have also been used to match implicit pose features, which can
be learned without requiring ground truth pose annotations~\cite{Sundermeyer2018-es}. 
The estimated 6D pose of the objects can be further refined~\cite{Rad2017-de,Li2018-fp} using an iterative procedure that effectively moves the camera around the object
so that the rendered image of the object best matches  %
the input image. Such a refinement step provides important performance improvements and is becoming common practice~\cite{Zakharov2019-fx,Wang2019-bb} as a final stage of the estimation
process. Our single-view single-object pose estimation described in Section~\ref{sec:single-view} builds on DeepIM
\cite{Li2018-fp}.
The performance of 6D pose estimation can be further improved using depth sensors~\cite{Li2018-fp,Xiang2018-dv,Wang2019-bb}, but in this work we focus on the most challenging scenario where only RGB images are available.

\paragraph{Multi-view single-object 6D pose estimation.}
Multiple views of an object can be used to resolve depth ambiguities and
gain robustness with respect to occlusions. Prior work using local invariant features includes~\cite{Collet2010-zj,Collet2011-lj,Grossberg2001-oh,Pless2003-ae} and involves some form of feature matching to establish correspondences across views to aggregate information from multiple viewpoints. 
More recently, the multi-view
single-object pose estimation problem has been revisited with a deep neural network
that predicts an object pose candidate in each view~\cite{Li2018-vl} and aggregates information from multiple views assuming known camera poses.
In contrast, our work does not assume the camera poses to be known.
We experimentally demonstrate that our approach outperforms~\cite{Li2018-vl} despite requiring less information.

\paragraph{Multi-view multi-object 6D pose estimation.} Other works consider all
objects in a scene together in order to jointly estimate the state of the scene
in the form of a compact representation of the object and camera poses in a common coordinate system. This problem is known as object-level
SLAM~\cite{Salas-Moreno2013-mt} where a depth-based object pose estimation method~\cite{Drost2010-zr} is used to recognize
objects from a database in individual images and estimate their poses. The individual objects are tracked across frames using depth measurements, assuming the motion of the sensor is continuous.
Consecutive depth measurements also enable to produce hypotheses for camera poses using ICP \cite{zhang1994iterative} and the poses of objects and cameras are finally refined in a joint optimization procedure.%
Another approach \cite{doumanoglou2016recovering} uses local RGBD patches to generate object hypotheses and find the best view of a scene. %
All of these methods, however, strongly rely on depth sensors to estimate the 3D structure of the scene while our method only exploits RGB images. In addition, they assume temporal continuity between the views, which is also not required by our approach.%

Other works have considered monocular RGB only object-level SLAM \cite{bao2011semantic,pillai2015monocular,yang2019cubeslam}.
  Related is also \cite{bachmann2019motion} where semantic 2D keypoint correspondences across multiple views and local features are used to jointly estimate the pose
  of a single human and the positions of the observing cameras. All of these works rely on local images features to estimate camera poses.
  In contrast, our work exploits 6D pose hypotheses generated by a neural
  network which allows to recover camera poses in situations where feature-based registration fails, as is the case for example for the complex texture-less images of the T-LESS dataset.
  In addition, \cite{pillai2015monocular,yang2019cubeslam} do not consider full 6D pose of objects, and
  \cite{bachmann2019motion,Li2018-vl} only consider scenes with a single instance of each object. In contrast, our method is able to handle scenes with multiple instances of the same object.

\section{Multi-view multi-object 6D object pose estimation}
\label{sec:method}

\begin{figure}[t]
  \centering
  \includegraphics[width=1.0\columnwidth]{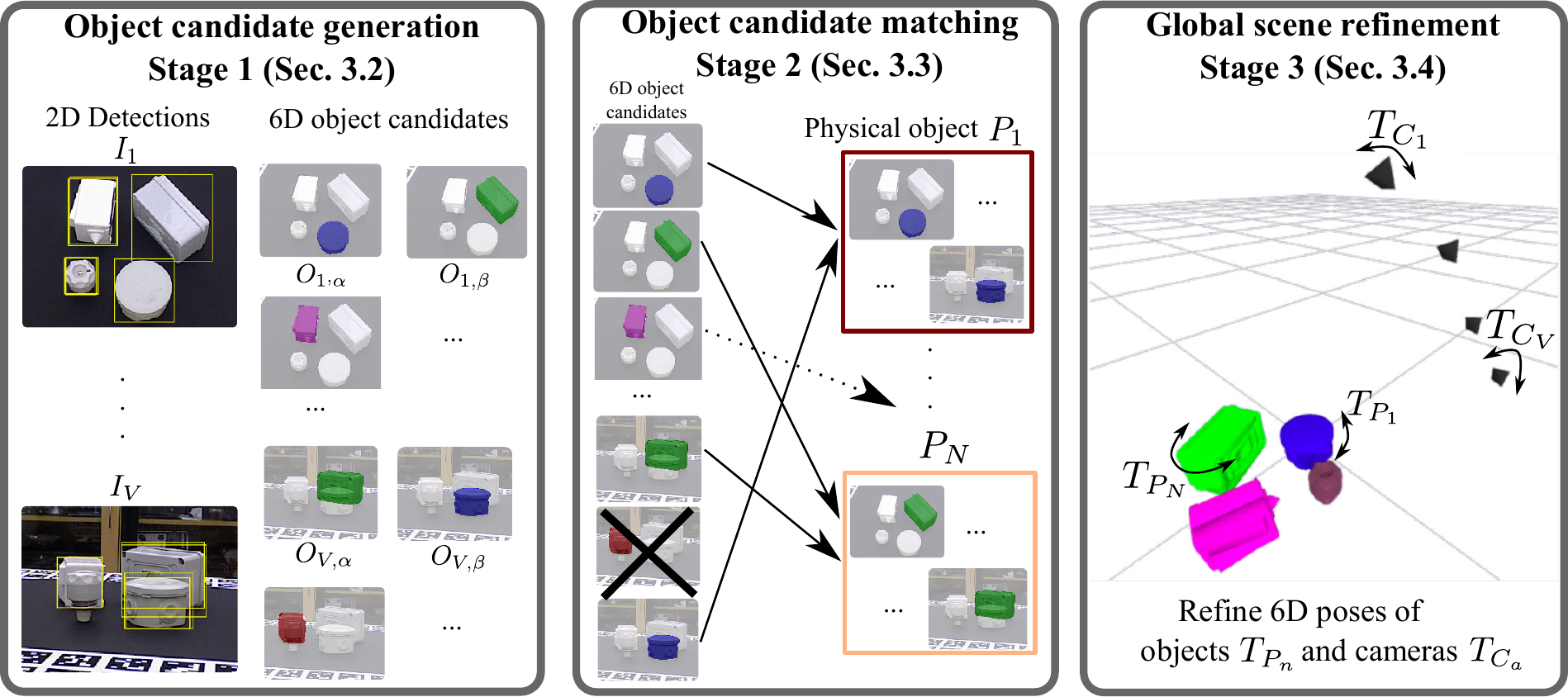}
  \caption{{\bf Multi-view multi-object 6D pose estimation.}  In the first stage, we obtain initial object candidates in each view separately. In the second stage, we match these object candidates across  views to recover a single consistent scene. In the third stage, we globally refine all object and camera poses to minimize multi-view reprojection error. %
  }
  \label{fig:method}
\end{figure}

In this section, we present our framework for multi-view multi-object pose estimation. We begin with an overview of the approach (Sec.~\ref{sec:method_overview} and Fig.~\ref{fig:method}), and then detail the three main steps of the approach in the remaining sections.  

\subsection{Approach overview}
\label{sec:method_overview}
Our goal is to reconstruct a scene composed of multiple objects given a set of
RGB images. We assume that we know the 3D models of objects of interest. 
However, there can be multiple objects of the same type in the scene and no information on the number or type of objects in the scene is available. Furthermore, objects may not be visible in some views, and the relative poses between the cameras are unknown.
Our output is a scene model, which includes the number of objects of each type, their 6D poses and the relative poses of the cameras. Our approach is composed of three main stages, summarized in Fig.~\ref{fig:method}.

In the first stage, we build on the success of recent methods for  single-view RGB object
detection and 6D pose estimation. Given a set of objects with known 3D models and a single image of a scene, we   output a set of candidate detections for each object and for each detection the 6D pose of the object with respect to the camera associated to the image. Note that some of these detections and poses are wrong, and some are missing. We thus consider the poses obtained in this stage as a set of initial {\it  object candidates}, i.e. objects that may be seen in the given view together with an estimate of their pose with respect to this view. This {\it object candidate generation} process is described in Sec.~\ref{sec:method_notations}.

In the second stage, called {\it object candidate matching} and described in detail in Sec.~\ref{sec:method_matching}, we match objects visible in multiple views to obtain a single consistent scene. 
This is a difficult problem since object candidates from the first stage typically include many errors due to  
(i) heavily occluded objects that might be mis-identified or for which the pose estimate might be completely wrong;
(ii) confusion between similar objects; and
(iii) unusual poses that do not appear in the training set and are not detected correctly. 
To tackle these challenges, we take inspiration from robust patch matching strategies that have been used in the structure from motion\,(SfM) literature~\cite{Szeliski_undated-fx,hartley2003multiple}. In particular, we design a matching strategy similar in spirit to~\cite{Rothganger2006-kh} but where we match entire 3D objects across views to obtain a single consistent 3D scene, rather than matching local 2D patches on a single 3D object~\cite{Rothganger2006-kh}.

The final stage of our approach, described in Section~\ref{sec:method_ba}, is a
global {\it scene refinement}. We draw inspiration from bundle adjustment~\cite{Triggs2000-gc}, but the optimization is performed at the level of objects: the 6D poses of all objects and cameras are refined to minimize a global reprojection error.

\subsection{Stage 1: object candidate generation}
\label{sec:method_notations}

Our system takes as input multiple photographs of a scene $\{I_a\}$ and a set of 3D models, each associated to an object label $l$. %
We assume the intrinsic
parameters of camera $C_a$ associated to image $I_a$ are known as is usually the case in single-view pose estimation methods.
In each view $I_a$, we obtain a set of object detections using an object detector (e.g. FasterRCNN\,\cite{Ren2017-ez}, 
RetinaNet\,\cite{Lin2017-wh}), and a set of candidate pose estimates  using a single-view single-object pose estimator (e.g. PoseCNN\,\cite{Xiang2018-dv}, DPOD\,\cite{Zakharov2019-fx}, DeepIM\,\cite{Li2018-fp}). 
While our approach is agnostic to the particular method used, we develop our own
single-view single-object pose estimator, inspired by DeepIM~\cite{Li2018-fp},
which improves significantly over state of the art and which we describe in the next paragraph. %
Each 2D candidate detection in view $I_a$ is identified by an index $\alpha$ and corresponds to an {\it object candidate} $O_{a,\alpha}$, associated with a predicted
object label $l_{a,\alpha}$ and a 6D pose estimate $\transform{C_a}{O_{a,\alpha}}$ with respect to camera $C_a$. 
We model a 6D pose
$\transform{}{} \in \SE$ as a $4\times4$ homogeneous matrix composed of a 3D rotation matrix
and a 3D translation vector. \\

{
\paragraph{Single-view 6D pose estimation.}
\label{sec:single-view}
We introduce a method for single-view 6D object pose estimation building on the idea of DeepIM~\cite{Li2018-fp} with some
simplifications and technical improvements.
First, we use a more recent neural-network architecture based on
EfficientNet-B3 \cite{Tan2019-my} and do not include auxiliary signals while
training. 
Second, we exploit the rotation parametrization
recently introduced in~\cite{Zhou2018-eg}, which has been shown to lead to more stable CNN training than quaternions. 
Third, we disentangle depth and translation prediction in the loss
following~\cite{Simonelli2019-da} and handle symmetries explicitly as
in~\cite{Wang2019-ml} instead of using the point-matching loss. 
Fourth, %
instead of fixing focal lengths to
1 during training as in~\cite{Li2018-fp}, we use focal lengths of the camera equivalent to the cropped images. 
Fifth, in addition to the real training images supplied with both
dataset, we also render a million images for each dataset using the provided CAD models for T-LESS
and the reconstructed models for YCB-Video. The CNNs are first pretrained using synthetic data only, then fine-tuned on both real
and synthetic images.
Finally, we use data augmentation on the RGB images while training our models,
which has been demonstrated to be crucial to obtain good performance on T-LESS~\cite{Sundermeyer2018-es}. 
We also note that this approach can be used for coarse estimation simply by providing a canonical pose as the input pose estimate during both training and testing. We rendered objects at a distance of $1$ meter from the camera and used this approach to perform coarse estimate on T-LESS. Additional details are provided in the appendix.
}

\paragraph{Object symmetries.} Handling object symmetries is a major challenge for object pose estimation since the object pose can only be estimated up to a symmetry. This is in particular true for our object candidates pose estimates. We thus need to consider symmetries explicitly together with the pose estimates. %
Each 3D model $l$ is associated to a set of symmetries $S(l)$. Following the framework introduced in
\cite{Pitteri2019-pt}, we define the set of symmetries  $S(l)$ as the set of transformations $S$ that leave the appearance of object $l$ unchanged:
\begin{equation}
  S(l) = \{S \in \mathrm{SE(3)} \: \mathrm{s.t} \: \forall\,\transform{}{} \in \mathrm{SE(3)}, \mathcal{R}(l, \transform{}{}) = \mathcal{R}(l, \transform{}{}S)\},
\end{equation}
where $\mathcal{R}(l,X)$ is the rendered image of object $l$ captured in pose
$X$ and $S$ is the rigid motion associated to the symmetry. 
Note
that $S(l)$ is infinite for objects that have axes of symmetry (e.g. bowls).

Given a set of symmetries $S(l)$ for the 3D object $l$, we define the symmetric distance $D_l$ which measures the distance
between two 6D poses represented by transformations $\transform{1}{}$ and $\transform{2}{}$.
Given an object $l$ associated to a set $\mathcal{X}_l$  %
of $|\mathcal{X}_l|$ 3D points $\mathrm{x}\in \mathcal{X}_l$, we define:
\begin{equation}
  \symdist{l}(\transform{1}{,} \transform{2}{}) =  \min_{S \in S(l)} \frac{1}{|\mathcal{X}_l|} \sum_{\mathrm{x} \in\mathcal{X}_l} || \transform{1}{} S \mathrm{x} - \transform{2}{} \mathrm{x}||_2.
  \label{eq:distance}
\end{equation}
$D_l(\transform{1}{},\transform{2}{})$ measures the average error
between the points transformed with $\transform{1}{}$ and $\transform{2}{}$
for the symmetry $S$ that best aligns the (transformed) points. In practice, to compute this distance for objects with axes of symmetries, we discretize $S(l)$ using $64$ rotation angles around each symmetry axis, similar to~\cite{Wang2019-ml}.

\subsection{Stage 2: object candidate matching}
\label{sec:method_matching}

As illustrated in Fig.~\ref{fig:method}, given the object candidates for all views $\{O_{a,\alpha}\}$, our matching module aims at (i) removing the object candidates that are not consistent across views and (ii) matching object candidates that correspond to the same
physical object. 
We solve this problem in two steps detailed below: (A) selection of candidate pairs of objects in all pairs of views, and (B) scene-level matching.

\paragraph{A. 2-view candidate pair selection.}
We first focus on a single pair of views $(I_a,I_b)$ of the scene and find all pairs of object candidates
$(O_{a,\alpha},O_{b,\beta})$, one in each view, which correspond to the same physical object in these two views.  %
To do so, we use a RANSAC procedure where we hypothesize a relative pose between the two cameras and count the number of inliers, i.e. the number of consistent pairs of object candidates in the two views.  We then select the solution with the most inliers which gives associations between the object candidates in the two views. %
In the rest of the section, we describe in more detail how we sample relative camera poses and how we define inlier candidate pairs. %

\paragraph{Sampling of relative camera poses.}

Sampling meaningful camera poses is one of the main challenges for our approach. 
Indeed, directly sampling at random the space of possible camera poses would be inefficient.
Instead, as usual in RANSAC, we sample pairs of object candidates (associated to the same object label) in the two views, hypothesize that they correspond to the same physical object and use them to infer a relative camera pose hypothesis. However, since objects can have symmetries, a single pair of candidates is not enough to obtain a relative pose hypothesis without ambiguities and we thus sample two pairs of object candidates, which in most cases is sufficient to disambiguate symmetries.

In detail, we sample two
tentative object candidate pairs with pair-wise consistent labels $(O_{a,\alpha}, O_{b,\beta})$ and
$(O_{a,\gamma},O_{b,\delta})$ and use them to build a relative camera pose hypothesis, $T_{C_aC_b}$. 
We obtain the relative camera pose hypothesis by (i) assuming that $(O_{a,\alpha}, O_{b,\beta})$ correspond to the same physical object and (ii) disambiguating symmetries by assuming that $(O_{a,\gamma},O_{b,\delta})$ also correspond to the same physical object, and thus selecting the symmetry that minimize their symmetric distance
\begin{eqnarray}
  \transform{C_a}{C_b}&& = \transform{C_a}{O_{a,\alpha}} S^\star \transform{C_b}{O_{b,\beta}}^{-1} \\
  &&~~ \mathrm{with} ~~
  S^\star = \argmin_{S \in S(l)}  \symdist{l}(\transform{C_a}{O_{a,\gamma}}, ( \transform{C_a}{O_{a,\alpha}}S \transform{C_b}{O_{b,\beta}}^{-1} )  \transform{C_b}{O_{b,\delta}} ) ,
\end{eqnarray}
\noindent where $l=l_{a,\alpha}=l_{b,\beta}$ is the object label associated to the first pair,  and $S^\star$ is the object symmetry which best aligns the point clouds associated to the second pair of objects $(O_{a,\gamma}$ and $O_{b,\delta})$. If the union of the two physical objects is symmetric, e.g. two spheres, the pose computed may be incorrect but it would not be verified by a third pair of objects, and the hypothesis would be discarded. %

\paragraph{Counting pairs of inlier candidates.}
Let's assume we are given a relative pose hypothesis between the cameras $\transform{C_a}{C_b}$. For each object candidate $O_{a,\alpha}$ in the first view, we  find the object candidate in the second view $O_{b,\beta}$ with the same label $l=l_{a,\alpha}=l_{b,\beta}$ that minimizes the symmetric distance $D_l(T_{C_aO_{a,\alpha}},T_{C_aC_b} T_{C_bO_{b,\beta}})$. In other words, $O_{b,\beta}$ is the object candidate in the second view closest to $O_{a,\alpha}$ under the hypothesized relative pose between the cameras. This pair $(O_{a,\alpha},O_{b,\beta})$ is considered an inlier if the associated symmetric distance is smaller than a given threshold $C$. The total number of inliers is used to score the relative camera pose $\transform{C_a}{C_b}$. Note that we discard the hypothesis which have fewer than three inliers. %

\paragraph{B. Scene-level matching.}
We use the result of the 2-view candidate pair selection applied to each image pair to define a graph between all candidate objects. %
Each vertex corresponds to an object candidate
in one view and edges correspond to pairs selected from 2-view candidate pair selection, i.e. pairs that had sufficient inlier support. We first remove isolated vertices, which correspond to object candidates that have not been validated by other views. %
Then, we associate to each connected component in the graph a unique physical object, which corresponds to a set of initial object candidates originating from different views. We call these physical objects $P_1,...P_N$ with $N$ the total number of physical objects, i.e. the number of connected components in the graph. We write $(a,\alpha)\in P_n$ to denote the fact that an object candidate $O_{a,\alpha}$ is in the connected component of object $P_n$. Since all the objects in a connected component share the same object label (they could not have been connected otherwise), 
we can associate without ambiguity an object label $l_n$ to each physical object $P_n$. 

\subsection{Stage 3: scene refinement}
\label{sec:method_ba}
  
After the previous stage, the correspondences between object candidates in the individual images are known, and the non-coherent object candidates have been removed. The final stage aims at recovering a unique and consistent scene model by performing global joint refinement of objects and camera poses.

In detail, the goal of this stage is to estimate poses of physical objects $P_n$,  represented by transformations $\transform{}{P_1}, \ldots, \transform{}{P_N}$, and cameras $C_v$, represented by transformations $\transform{}{C_1}, \ldots, \transform{}{C_V}$, in a common world coordinate frame. 
This is similar to the standard bundle adjustment problem where the goal is to recover the 3D points of a scene together with the camera poses. This is typically addressed by minimizing a reconstruction loss that measures the 2D discrepancies between the projection of the 3D points and their measurements in the cameras. In our case, instead of working at the level of points as  done in the bundle adjustment setting, we introduce a reconstruction loss that operates at the level of objects. 

More formally, for each object present in the scene, we introduce an object-candidate reprojection loss accounting for symmetries. %
We define the loss for a candidate object $O_{a,\alpha}$ associated to a physical object $P_n$ (i.e. $(a,\alpha)\in P_n$) and the estimated candidate object pose $\transform{C_a}{O_{a,\alpha}}$ with respect to $C_a$ as: %
\begin{multline}
  L\left(\transform{}{P_n},\transform{}{C_a}|\transform{C_a}{O_{a,\alpha}}\right)=%
  \min_{S \in S(l)} \frac{1}{|\mathcal{X}_{l}|}\sum_{\mathrm{x}\in \mathcal{X}_{l}} ||\pi_{a}(\transform{C_a}{O_{a,\alpha}} S \mathrm{x}) - \pi_{a}(\transform{}{C_a}^{-1}\transform{}{P_n}\mathrm{x})||,
  \label{eq:reprojection_loss}
\end{multline}
where $|| \cdot ||$ is a truncated L2 loss, $l=l_n$ is the label of the physical object $P_n$,  $\transform{}{P_n}$ the 6D pose of object $P_n$ in the world coordinate frame, $\transform{}{C_a}$ the pose of camera $C_a$  in the world coordinate frame, $\mathcal{X}_{l}$ the set of 3D points associated to the 3D model of object $l$, $S(l)$ the symmetries of the object model $l$, and the operator $\pi_{a}$ corresponds to the 2D projection of
3D points expressed in the camera frame $C_a$ by the intrinsic calibration matrix of camera $C_a$.
The inner sum in Eq.~\eqref{eq:reprojection_loss} is the error between (i) the 3D points x of the object model $l$ projected to the image with the single view estimate of the transformation $\transform{C_a}{O_\alpha}$ that is associated with the physical object (i.e. $(a, \alpha)\in P_n$) (first term, the image measurement) and (ii) the 3D points $\transform{P_n}{}\mathrm{x}$ on the object $P_n$ projected to the image by the global estimate of camera $C_a$ (second term, global estimates). %

Recovering the state of the unique scene which best explains the measurements consists in solving the following consensus optimization problem:
\begin{equation}
  \min_{\transform{}{P_1}, \ldots, \transform{}{P_N} , \transform{}{C_1}, \ldots, \transform{}{C_V}} \sum_{n=1}^N \sum_{(a,\alpha)\in P_n} \,L\left(\transform{}{P_n},\transform{}{C_a}|\transform{C_a}{O_{a,\alpha}}\right),
   \label{eq:ba_problem}
\end{equation}
where the first sum is over all the physical objects $P_n$ and the second one over all object candidates $O_{a,\alpha}$ corresponding to the physical object $P_n$. In other words, we wish to find global estimates of object poses  $\transform{}{P_n}$ and camera poses $\transform{}{C_a}$ to match the (inlier) object candidate poses $\transform{C_a}{O_{a,\alpha}}$ obtained in the individual views.
{The optimization problem is solved using the
Levenberg-Marquart algorithm.
We provide more details in the appendix. 
}

\section{Results}
\label{sec:experiments}

In this section, we experimentally evaluate our method on the \mbox{YCB-Video} \cite{Xiang2018-dv}
 and \mbox{T-LESS} \cite{Hodan2017-pq} datasets, which both provide multiple views and ground truth 6D object poses for cluttered scenes with multiple objects.  
In Sec.~\ref{sec:exp_singleview}, we first validate and analyze our single-view single-object 6D
pose estimator. 
We notably show that our single-view single-object 6D pose estimation method already
improves state-of-the-art results on both datasets. %
In Sec.~\ref{sec:exp_multiview}, we validate our multi-view multi-object framework by demonstrating consistent improvements over the
single-view baseline. %

\setlength{\tabcolsep}{4pt}
\begin{table}[t]
    \caption{{\bf Single-view 6D pose estimation. } Comparisons with state-of-the-art methods on the YCB-Video (a) and T-LESS datasets (b).}%
  \label{tab:singleview}
  \begin{center}
	\begin{minipage}{0.52\textwidth}
  \begin{center}
    \begin{tabular}{l|cc}
    & AUC of  & AUC of\\
    & ADD-S & ADD(-S)\\
    \hline
      PoseCNN \cite{Xiang2018-dv} & -             & 61.3   \\
      MCN \cite{Li2018-vl}        & 75.1          & -  \\
      PVNet \cite{Peng2018-ev}    & -             & 73.4  \\
      DeepIM \cite{Li2018-fp}     & 88.1          & 81.9 \\
      Ours                        & \textbf{89.8} & \textbf{84.5}
    \end{tabular}
    \caption*{(a) YCB-Video}
    \end{center}
    \end{minipage}
	\begin{minipage}{0.47\textwidth}
    \begin{center}
      \label{tab:singleview_tless}
    \begin{tabular}{l|c}
    &$e_{\mathrm{vsd}}<0.3$\\
    \hline
        Implicit\,\cite{Sundermeyer2018-es} & 26.8 \\
        Pix2pose\,\cite{Park2019-od} & 29.5  \\
        Ours & \textbf{63.8}\\ %
        ~~w/o loss & 60.1 \\
        ~~w/o network& 59.5 \\
        ~~w/o rot.& 61.0 \\
        ~~w/o data augm.& 37.0 \\
    \end{tabular}
    \caption*{(b) T-LESS SiSo task}
  \end{center}
    \end{minipage}
  \end{center}
\end{table}

\subsection{Single-view single-object experiments}
\label{sec:exp_singleview}

\paragraph{Evaluation on YCB-Video.} Following~\cite{Xiang2018-dv,Li2018-fp,Peng2018-ev}, we evaluate on a subset of 2949
keyframes from videos of the 12 testing scenes. 
We use the standard
ADD-S and ADD(-S) metrics and their area-under-the-curves~\cite{Xiang2018-dv}
(please see appendix for details on the metrics). We evaluate
our refinement method using the same detections and
coarse estimates as DeepIM~\cite{Li2018-fp}, provided by
PoseCNN~\cite{Xiang2018-dv}. We ran two iterations of pose refinement network.
Results are shown in Table~\ref{tab:singleview}a. 
Our method improves over the current-state-of-the-art DeepIM~\cite{Li2018-fp}, by approximately 2 points on the AUC of ADD-S and ADD(-S) metrics. %

\paragraph{Evaluation on T-LESS.}
As explained {in Section~\ref{sec:single-view}}, we use our single-view approach both for coarse pose estimation and refinement. %
We compare our method against the two recent RGB-only methods
Pix2Pose\,\cite{Park2019-od} and Implicit\,\cite{Sundermeyer2018-es}. For a fair
comparison, we use the detections from the same RetinaNet model as in\,\cite{Park2019-od}. 
We report results on the SiSo task \cite{Hodan_undated-sl}
and use the standard visual surface discrepancy (vsd) recall metric with the same parameters as in~\cite{Park2019-od,Sundermeyer2018-es}.
Results are presented in Table~\ref{tab:singleview}b.  
On the $e_{\mathrm{vsd}}<0.3$ metric, our \{coarse\,+\,refinement\} solution achieves a significant {$34.2\%$} absolute
improvement compared to existing state-of-the-art methods.
Note that~\cite{Li2018-fp} did not report results on
T-LESS.
We also evaluate on this dataset the benefits of the key components of our single view approach compared to the components used in DeepIM\cite{Li2018-fp}. 
  More precisely, we evaluate the
  importance of the base network (our EfficientNet vs FlowNet pre-trained), loss (our
  symmetric and disentangled vs. point-matching loss with $L_1$ norm), rotation parametrization (our using
  \cite{Zhou2018-eg} vs. quaternions) and data augmentation (our color
  augmentation, similar to \cite{Sundermeyer2018-es} vs. none). Loss, network and rotation parametrization bring a small but clear improvement.
  Using data augmentation is crucial on the T-LESS dataset where training is performed only on synthetic data and real images of the objects on dark background.

\subsection{Multi-view experiments}
\label{sec:exp_multiview}

As shown above, our single-view method achieves state-of-the-art results on both datasets. 
We now evaluate the performance of our multi-view approach to estimate 6D poses in scenes with multiple objects and multiples views.

\paragraph{Implementation details.} On both datasets, we use the same hyper-parameters. In stage 1, we only consider object detections with a score superior to 0.3 to limit the number of detections. In stage 2, we use a RANSAC 3D inlier threshold of $C=2\,$cm.
This low threshold ensures that no outliers are considered while associating object candidates. 
We use a maximum number of {$2000$} RANSAC iterations for each pair of views, but this limit is only reached for the
most complex scenes of the T-LESS dataset containing tens of detections. 
For instance, in the context of two views with six different 6D object
candidates in each view, only 15 RANSAC iterations are enough to explore all relative camera pose hypotheses. 
For the scene refinement (stage 3), we use 100 iterations of Levenberg-Marquart (the optimization
typically converges in less than 10 iterations).

\paragraph{Evaluation details.} In the single-view evaluation, the poses of the objects are expressed with respect to the camera frame. To fairly compare with the single-view baseline, we also
evaluate the object poses in the camera frames, that we compute using the absolute object poses
and camera placements estimated by our global scene refinement method. 
Standard metrics for 6D pose estimation strongly penalize methods with low detection recall. 
To avoid being penalized for removing objects that cannot be verified across several views, we thus add the initial object
candidates to the set of predictions but with
confidence scores strictly lower than the predictions from our full scene
reconstruction. 

\paragraph{Multi-view multi-object quantitative results.} The problem that we consider,
recovering the 6D object poses of multiple known objects in a scene captured by several RGB images taken from unknown viewpoints
has not, to the best of our knowledge, been addressed by prior work reporting results on the YCB-Video and T-LESS datasets. 
The closest work is
\cite{Li2018-vl}, which considers multi-view scenarios on YCB-Video and uses
ground truth camera poses to align the viewpoints. In~\cite{Li2018-vl}, results are provided for
prediction using 5 views. We use our approach with the same number of input images but without using ground truth calibration and report results in Table~\ref{tab:multiview}a. Our method significantly outperforms
\cite{Li2018-vl} in both single-view and multi-view scenarios.

\setlength{\tabcolsep}{4pt}
\begin{table}[t]
    \caption{{\bf Multi-view multi-object results.} (a) Our approach significantly
      outperforms~\cite{Li2018-vl} on the YCB-Video dataset in both the single view and multi-view scenarios while not requiring
    known camera poses. (b) Results on the T-LESS dataset. Using multiple views clearly improves our results.}
    \label{tab:multiview}
	\begin{minipage}{0.34\textwidth}
  \begin{center}
    \begin{tabular}{c|cc}
        & 1 view & 5 views \\
      \hline
      \cite{Li2018-vl}    & 75.1 & 80.2 \\
      Ours & 89.8  & \textbf{93.4}   \\
    \end{tabular}
      \caption*{\centering (a) YCB-Video\newline(AUC of ADD-S)}
  \end{center}
    \end{minipage}
	\begin{minipage}{0.65\textwidth}
     \begin{center}
      \begin{tabular}{c|ccc}
        & 1 view  &  4 views &  8 views    \\
        \hline
        \multicolumn{1}{r|}{AUC of ADD-S}         & 72.1    &  76.0       & \textbf{78.9}  \\
        \multicolumn{1}{r|}{ADD-S $<$ 0.1d}       & 68.0    &  72.6       & \textbf{76.6}  \\
        \multicolumn{1}{r|}{$e_{\mathrm{vsd}}<0.3$} & 62.6    &  67.6       & \textbf{71.6}  \\
        \multicolumn{1}{r|}{mAP@ADD-S$<$0.1d}     & 55.0    &  61.6       & \textbf{69.0}  \\
      \end{tabular}
      \caption*{(b) T-LESS ViVo task (ours, 1000 images)}
    \end{center}
    \end{minipage}
\end{table}

We also perform multi-view experiments on T-LESS with a variable number of
views. {We follow the multi-instance
  BOP\cite{Hodan_undated-sl} protocol for ADD-S$<$0.1d and
  $e_{\mathrm{vsd}}<0.3$. We also analyze precision-recall
  tradeoff similar to the standard practice in object detection.
  We consider positive predictions that satisfy ADD-S$<$0.1d and report
  mAP@ADD-S$<$0.1d. Results are shown in Table~\ref{tab:multiview}b for the ViVo
  task on 1000 images. To the best of our knowledge, no other method has reported results on this task. As expected, our multi-view approach brings significant improvements compared to only single-view baseline. }

\setlength{\tabcolsep}{4pt}
\begin{table}[t]
    \caption{Benefits of the scene refinement stage. We report pose
      ADD-S errors (in mm) for the inlier object
      candidates before and after global scene refinement. Scene-refinement
      improves 6D pose estimation accuracy.}
    \label{tab:xyz_errors}
  \begin{center}
    \begin{tabular}{c|cc}
      \multicolumn{1}{c|}{} & \multicolumn{1}{c}{YCB dataset} & \multicolumn{1}{c}{T-LESS dataset}\\
      \hline
        Before refinement   & 6.40          & 4.43 \\
        After refinement    & \textbf{5.05} & \textbf{3.19} \\
      \hline
    \end{tabular}
  \end{center}
\end{table}

\paragraph{Benefits of scene refinement.} To demonstrate the benefits
of global scene refinement (stage 3), we report in Table~\ref{tab:xyz_errors}
the average ADD-S errors of the inlier candidates before and after solving the
optimization problem of Eq.\eqref{eq:ba_problem}.
{We note a clear relative improvement, around 20\% on both datasets.}.

\paragraph{Relative camera pose estimation.}
A key feature of our method is
  that it does not require camera position to be known and instead robustly estimates it
  from the 6D object candidates. We investigated alternatives to our joint camera
  pose estimation. First, we used COLMAP\cite{schoenberger2016sfm,schoenberger2016mvs}, a popular
  feature-based SfM software, to recover camera poses. On randomly sampled
  groups of 5 views from the YCB-Video dataset COLMAP outputs camera poses in
  only $67\%$ of cases compared to $95\%$ for our method. On groups of 8 views
  from the more difficult T-LESS dataset, COLMAP outputs camera poses only in
  4\% of cases, compared to 74\% for our method. Our method therefore demonstrates a
  significant interest compared to COLMAP that uses features to recover
  camera poses, especially for complex textureless scenes like in the T-LESS
  dataset. Second, instead of estimating camera poses using our approach, we
  investigated using ground truth camera poses available for the two datasets.
  We found that the improvements using ground truth camera poses over the camera poses
  recovered automatically by our method were only minor: within $1\%$ for T-LESS (4 views) and YCB-Video (5 views),
  and within $3\%$ for T-LESS (8 views). {This demonstrates that our approach recovers accurate camera poses
  even for scenes containing only symmetric objects as in the T-LESS dataset.}

  \paragraph{Qualitative results.} We provide examples of recovered 6D object poses in Fig.~\ref{fig:visualization} where we show both object candidates and the final estimated scenes.
{\bf Please see the appendix for additional results}, including detailed
discussion of failure modes.
Results on the YCB-Video are available \href{https://www.di.ens.fr/willow/research/cosypose/}{on the project webpage}\footnote{\href{https://www.di.ens.fr/willow/research/cosypose/}{https://www.di.ens.fr/willow/research/cosypose/}}.

\begin{figure}
  \centering
  \includegraphics[width=1.0\columnwidth]{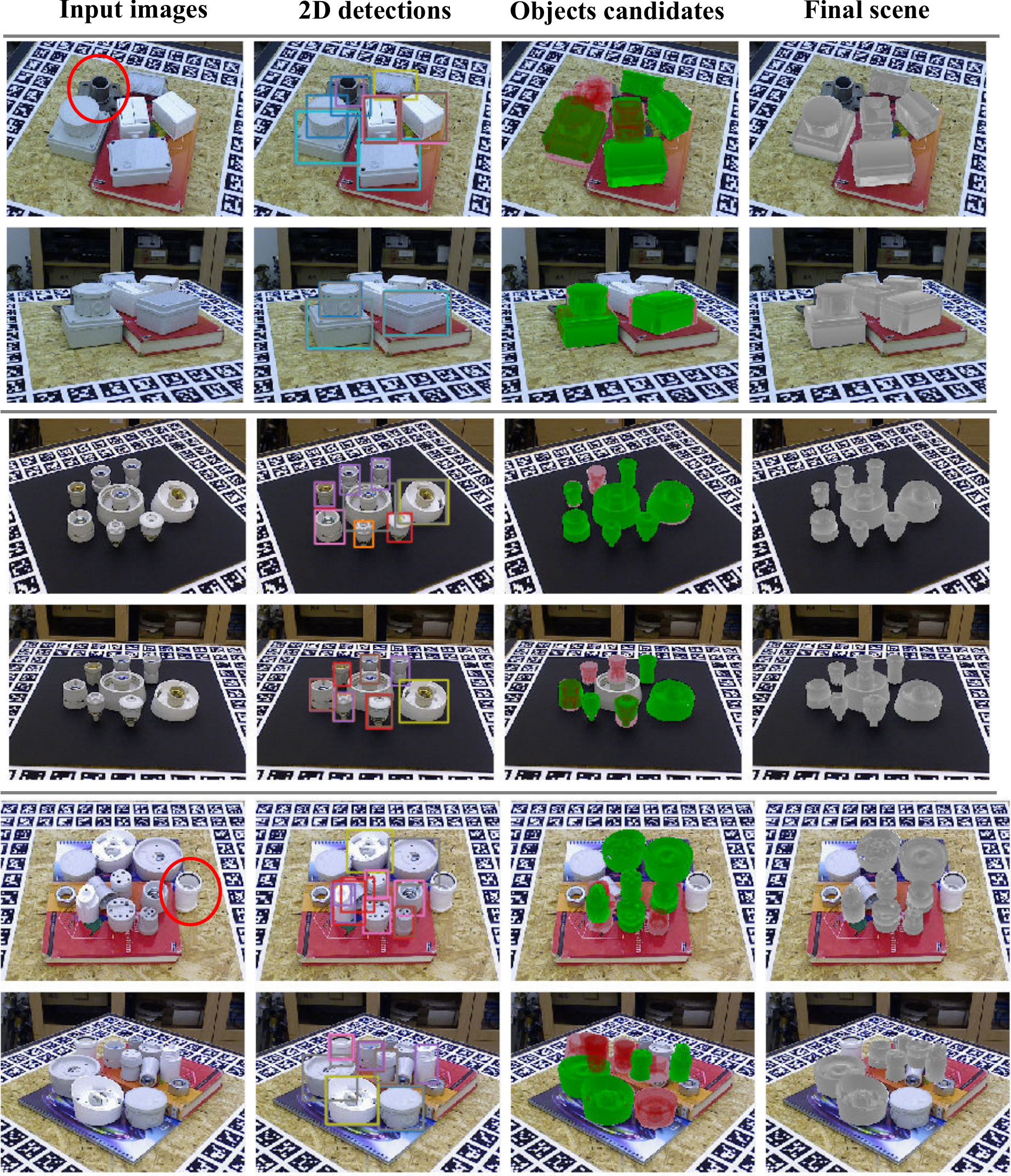}
  \caption{{\bf Qualitative results.} We present three examples of scene
    reconstructions. For each scene, two (out of 4) views that were used to reconstruct the scene are shown as two rows. 
    In each row, the first column shows the input RGB image. The second column shows the 2D detections. The third column shows all object candidates with marked inliers (green) and outliers (red).
    The fourth column shows the final scene reconstruction. Objects marked by red circles are not in the database, but are sometimes incorrectly detected.
    Notice how our method estimates accurate 6D object poses for many objects in challenging scenes containing texture-less and symmetric objects, severe occlusions, and where many objects are similar to each other. {\bf More examples are in the appendix.}}
  \label{fig:visualization}
\end{figure}

\paragraph{Computational cost.}
{For a common case with 4 views and 6 2D detections per view, our approach takes
  approximately 320\,ms to predict the state of the scene. This timing includes: 190\,ms 
  for estimating the 6D poses of all candidates (stage 1, 1 iteration of the coarse and refinement networks), 40\,ms for the object
  candidate association (stage 2) and 90\,ms for the scene refinement (stage 3). Further speed-ups towards real-time performance could be achieved, for example, by exploiting temporal continuity in a video sequence.}

\section{Conclusion}
\label{sec:conclusion}

We have developed an approach, dubbed CosyPose, for recovering the 6D pose of multiple known objects viewed by several non-calibrated cameras. Our main contribution is to combine learnable 6D pose estimation with robust multi-view matching and global refinement to reconstruct a single consistent scene. Our approach explicitly handles object symmetries, does not require depth measurements, is robust to missing and incorrect object hypothesis, and automatically recovers the camera poses and  the number of  objects  in  the  scene. 
These results make a step towards the robustness and accuracy
required for visually driven robotic manipulation in unconstrained scenarios with moving cameras, and open-up the possibility of including object pose estimation in an active visual perception loop.

\section*{Acknowledgments} This work was partially supported by the HPC resources from GENCI-IDRIS (Grant 011011181), the European Regional Development Fund under the project IMPACT (reg. no. CZ.02.1.01/0.0/0.0/15 003/0000468), Louis Vuitton ENS Chair on Artificial Intelligence, and the French government under management of Agence Nationale de la Recherche as part of the "Investissements d'avenir" program, reference ANR-19-P3IA-0001 (PRAIRIE 3IA Institute).

\bibliographystyle{splncs}
\bibliography{biblio}

\appendix
\section*{Appendix}
The appendix is organized as follows. In Sec.~\ref{sec:singleview}, we give more details of our single-view
single-object 6D object pose estimator. In Sec.~\ref{sec:supmat-matching} we illustrate the object candidate matching
strategy on a simple 2D example. In Sec.~\ref{sec:optim}, we give additional details about our parametrization and initialization of the object-level bundle adjustment problem, introduced in Sec.~\ref{sec:method_ba} of the main paper. Sec.~\ref{sec:datasetmetric} presents the datasets used in the main
paper and recalls the metrics that are used for each dataset.
Finally, in Sec.~\ref{sec:qualitative_examples} we present additional qualitative results of our
multi-view multi-object 6D pose estimation approach. We discuss in detail some
examples to illustrate key benefits of our method as well as point out the main
limitations. Examples randomly selected from the results on the T-LESS and
YCB-Video datasets are available \href{https://www.di.ens.fr/willow/research/cosypose/}{on the project webpage}\footnote{\href{https://www.di.ens.fr/willow/research/cosypose/}{https://www.di.ens.fr/willow/research/cosypose/}}.

\section{Our single-view single-object method}
\label{sec:singleview}
We now detail our single-view single-object pose estimation network 
introduced in Sec.~\ref{sec:method_notations} of the main paper. Our method builds on DeepIM \cite{Li2018-fp}
but includes several extensions and improvements.

Given a single image $I_a$ and a 2D detection $D_{a,\alpha}$ associated with
an object label $l_{a,\alpha}$, our method outputs an hypothesis for the pose
of the object with respect to the camera. This pose is noted $T_{C_a,O_{a\alpha}}$. In this
section, we focus on one view and one object and thus omit the $a$ and $\alpha$ subscripts. 
Similar to DeepIM \cite{Li2018-fp}, we use a deep neural network that takes as input
two images and iteratively refines the pose. The first image is the (real) input image $I$ cropped on a region of the image showing the
object, denoted $I^c$. At iteration $k$, the second image is a (synthetic) rendering of the object with label $l$
rendered in a pose $T_{C,O}^{k-1}$ that corresponds to the object pose estimated at the previous iteration. The network outputs
an updated refined pose $T_{C,O}^{k}$. 
The initial pose $T_{C,O}^0$ can be provided by any coarse 6D pose estimation
method (such as PoseCNN~\cite{Xiang2018-dv}) but we also show that we can simply
use a canonical pose of the object for $T_{C,O}^0$ as explained in the ``Coarse
estimation'' pagraph below. We now detail our method and present the main
differences with \cite{Li2018-fp}.

\subsubsection{Network architecture.}
The network takes as input the concatenation of the synthetic and real cropped
images. Both images are resized to the input resolution: $320\times240$.
The backbone is EfficientNet-B3 \cite{Zhou2018-eg} followed by spatial average
pooling. The prediction layer is a simple fully connected layer which outputs 9 values corresponding to one vector
$[ v_x, v_y, v_z ]$ for the translation and two vectors $e_{1},e_{2}$ to predict the
rotation component of $\transform{C}{O}$. A rotation matrix $R$
is recovered from $e_{1},e_{2}$ using \cite{Zhou2018-eg} by simply
orthogonalizing the basis defined by the two vectors $e_{1},e_{2}$. Please see
``Rotation parametrization'' for the equations to recover the rotation matrix
$R$ from $e_{1},e_{2}$. Compared to DeepIM \cite{Li2018-fp}, the main difference is
that we use a more recent network architecture (DeepIM is based on FlowNet
\cite{dosovitskiy2015flownet}) and we do not include auxiliary predictions of flow
and mask. This makes the method simpler and easier to train. Our input resolution of
$320\times240$ is also smaller than $640\times480$ used by DeepIM, reducing
memory consumption and allowing to use larger batches while training. 

\subsubsection{Transformation parametrization.}Similar to DeepIM, we use the object-independent
rotation and translation parametrization which consists in predicting a rotation of the camera around the object,
a $xy$ translation $[v_x, v_y]$ in image space (in pixels) for the center of the
rendered object and a relative displacement $v_z$ along the depth axis of the camera.
Given the input pose $T_{CO}^k$ and the outputs of the network ($[v_x, v_y,
v_z]$ and $R=f(e_1,e_2)$), the pose update is obtained from the following equations:

\begin{eqnarray}
  x^{k+1} & = & \left(\frac{v_x}{f_x^C} + \frac{x^{k}}{z^{k}}\right) z^{k+1} \label{eq:x}\\
  y^{k+1} & = & \left(\frac{v_y}{f_y^C} + \frac{y^{k}}{z^{k}}\right) z^{k+1} \label{eq:y}\\
  z^{k+1} & = &  v_z z^{k} \label{eq:depth_param}\\
  R^{k+1} & = & R R^{k} \label{eq:R},
\end{eqnarray}

\noindent where $[x^{k}, y^{k}, z^{k}]$ is the 3D translation vector of
$T_{CO}^{k}$, $R^k$ the rotation matrix of $T_{CO}^{k}$, $f_x^C$ and $f_y^C$ are
the focal lengths that correspond to the (fictive) camera associated with the
\textbf{cropped} input image $I^C$. Finally,  $[x^{k+1}, y^{k+1}, z^{k+1}]$ and
$R^{k+1}$ are the parameters of the output pose estimate $T_{CO}^{k+1}$. The
differences with DeepIM are twofold. First, we use a linear parametrization of
the relative depth (eq.~\eqref{eq:depth_param}), instead of
$z^{k+1}=z^{k}e^{-v_z}$, which we found more stable to train. Second, we use the
intrinsics $f_x^C$, $f_y^C$ of the cropped camera associated with the input
(cropped) image. DeepIM uses the intrinsics parameters of the non-cropped camera
$f_x$, $f_y$ and fix them to $1$ during training because the intrinsic parameters
of the input camera are fixed on their datasets. We use the cropped focal lengths
instead because (a) cropping and resizing the crop of the input image changes the apparent focal length and (b)
the focal lengths of the input images are not unique on T-LESS. Using the
cropped focal lengths forces the network to only predict $xy$ translations in
pixels and the network can therefore become invariant to the intrinsic parameters of the input
(cropped) camera.

\subsubsection{Rotation parametrization.} Given two vectors $e_1$ and $e_2$ (6
values) predicted by the neural network, we recover a rotation parametrization $R$
by following \cite{Zhou2018-eg}:
\begin{eqnarray}
  e_1^\prime & = & \frac{e_1}{||e_1||_2} \\
  e_3^\prime & = & \frac{e_1^\prime \wedge e_2}{||e_2||_2} \\
  e_2^\prime & = & e_3^\prime \wedge e_1^\prime,
\end{eqnarray}

\noindent where $\wedge$ is the cross product between two 3D vectors. This
representation has been shown to be better than quaternions (used by DeepIM) to
regress with a neural network \cite{Zhou2018-eg}.

\subsubsection{Cropping strategy.}  DeepIM uses (a) the input 2D detections and
(b) the bounding box defined by $T_{CO}^{k}$ and the vertices of the object $l$ to define the size and location
of the crop in the real input image during training. Indeed, the ground truth
bounding box is known during training. At test time, only (b) is used by DeepIM because ground
truth bounding boxes are not available. In our case, we only use
(b) while training and testing. The intrinsic parameters of the cropped camera
are also used to directly render the cropped synthetic image at
a resolution of $320\times240$ instead of rendering at a larger resolution followed by cropping.

\subsubsection{Symmetric disentangled loss.} A standard loss for 6D pose estimation is
ADD-S\cite{Xiang2018-dv} which allows to predict pose of symmetric objects. Our loss is inspired by ADD-S loss with two main differences.
First, we enumerate all the possible symmetries to find the best matching between the vertices of the
predicted model and the ground truth model instead of finding the nearest
neighbors. This is similar in spirit to the approach of~\cite{Wang2019-ml} to handle object
symmetries. Second, we disentangle depth $v_z$ and translation predictions
$v_x,v_y$, following the recommendations from
\cite{Simonelli2019-da}.

More formally, we define the update function $F$ which takes as input the initial estimate
of the pose $T_{CO}^k$, the outputs of the neural network $[v_x, v_y, v_z]$ and
$R$, and outputs the updated pose, i.e. the function such that
\begin{equation}
  \centering
T_{CO}^{k+1} = F(T_{CO}^k, [v_x, v_y, v_z], R),
\end{equation}
\noindent where the closed form of $F$ is expressed in
equations~\eqref{eq:x}\eqref{eq:y}\eqref{eq:depth_param}\eqref{eq:R} of the
appendix. 
We also write $[\hat{v}_x, \hat{v}_y, \hat{v}_z]$ and $\hat{R}$ the target
predictions,  i.e. the predictions such that $\hat{T}_{CO} = F(T_{CO}^k,
[\hat{v}_x, \hat{v}_y, \hat{v}_z], \hat{R})$, where $\hat{T}_{CO}$ is the ground
truth pose of the object. 
Our loss function is then:
\begin{align}
\mathcal{L}(T_{CO}^k,[v_x, v_y, v_z], R) & = D_l(F(T_{CO}^k, [v_x,v_y,\hat{v}_z], \hat{R}), \hat{T}_{CO}) \label{eq:loss-xy}\\
                                       &\hspace*{0.2cm} + D_l(F(T_{CO}^k, [\hat{v}_x,\hat{v}_y,v_z], \hat{R}), \hat{T}_{CO})\label{eq:loss-z}  \\
                                       &\hspace*{0.2cm} + D_l(F(T_{CO}^k, [\hat{v}_x,\hat{v}_y,\hat{v}_z], R), \hat{T}_{CO})\label{eq:loss-R},
\end{align}
\noindent where $D_l$ is the symmetric distance defined in the Sec.~\ref{sec:method_notations} of the
main paper{, with the $L_2$ norm replaced by the $L_1$ norm.} The different terms of this loss separate the influence of:
$xy$ translation~\eqref{eq:loss-xy}, relative depth~\eqref{eq:loss-z} and rotation~\eqref{eq:loss-R}. We refer to
\cite{Simonelli2019-da} for additional explanations of the loss disentanglement.

\subsubsection{Coarse estimation.} To perform coarse estimation on T-LESS, we
use the same network architecture, parametrization and losses defined above. As
input $T_{CO}^{0}$ we provide a canonical input pose that corresponds to the
object being rendered at a distance of $1$ meter of the camera in the center of
the input 2D bounding box. The coarse and refinement networks use the same
architecture, but the weights are distinct. Each network is trained independently.

\subsubsection{Training data.} Due to the complexity of annotating real data with 6D
pose at large scale, most recent methods
\cite{Zakharov2019-fx,Li2018-fp,Sundermeyer2018-es} generate additionnal
synthetic training data. %
In our experiments, we use the real training images provided by YCB-Video and
the images of the real objects displayed individually on black backgrounds provided by T-LESS. In addition, we
generate one million synthetic training images on each dataset using a simple
procedure described next.

We randomly sample 3 to 9 objects from the set of 3D models
considered, place them randomly in a 3D box of size 50 cm and sample randomly the orientation of each
object. Half of the images are generated with objects flying
in the air, the other half is generated by taking the images after running physics simulation for a
few seconds, generating physically feasible object configurations. This is similar to the approach described in ~\cite{Tremblay2018-bd,Tremblay2018-lq}, though none of our rendered images are
photorealistic. The camera is pointed at the center of the 3D box, its position is
sampled uniformly above the box center at the same range of distance as the one of the real training data, and its roll angle is sampled between (-10, 10) degrees. 
On T-LESS, the distance to the object is fixed in the real training images and we use instead the range of distances of the testing set provided (which is explicitly allowed by the guidelines of the
BOP challenge \cite{Hodan_undated-sl}\footnote{See
  https://bop.felk.cvut.cz/challenges/ Sec 2.2.}). We do not use any information from the
testing set beside this distance interval.

On the T-LESS dataset, we generate data using the CAD models only.
We add random textures on the CAD models following work on domain
randomization \cite{Tobin2017-mc,Loing2018-ft,labbe2020monte}. 
We also paste images from the Pascal VOC dataset in the background with a probability 0.3, following \cite{Li2018-fp}.
On both datasets, we add data augmentation to the input RGB
images while training, following \cite{Sundermeyer2018-es}. Data augmentation includes gaussian blur, contrast,
brightness, color and sharpness filters from the Pillow library \cite{pillow}.

\begin{figure}
  \centering
  \includegraphics[width=1.0\columnwidth]{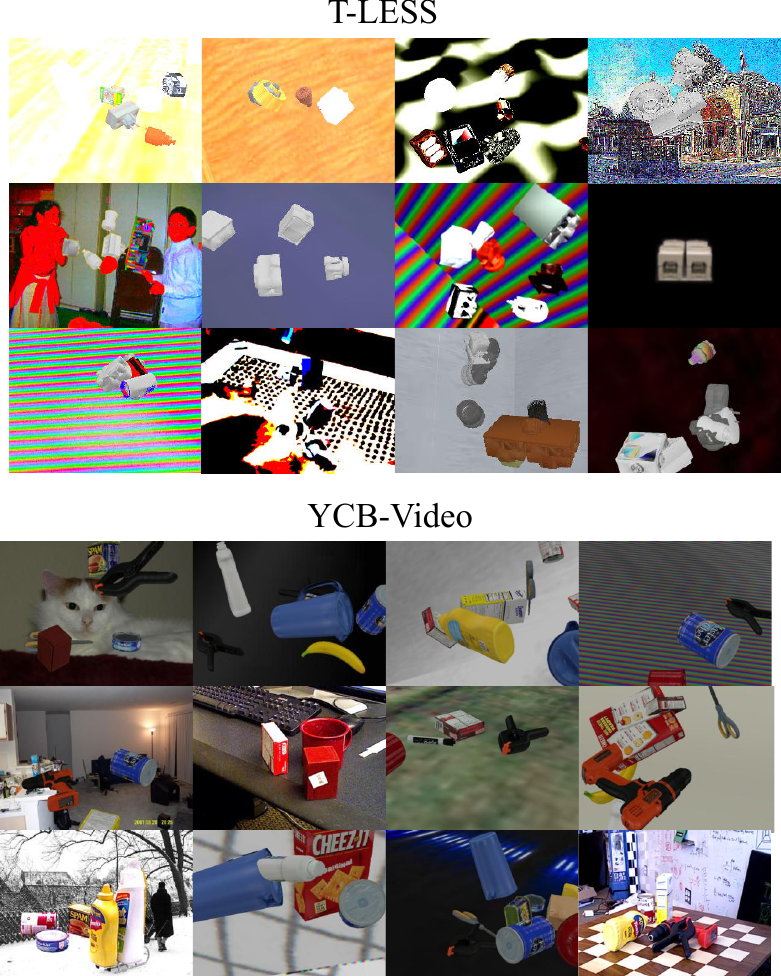}
  \caption{\small {\bf Training images for our single-view single-object pose
      estimation networks.} Examples of training images used for
    training the networks on T-LESS and YCB-Video.}
  \label{fig:training_images}
\end{figure}

Examples of training images are shown in Fig.~\ref{fig:training_images}.
Finally, when training the refinement network, we use the same distribution as DeepIM for the input poses.

\subsubsection{Training procedure.}
All of the networks (refinement network on YCB-Video, coarse network on T-LESS,
refinement network on T-LESS) are trained using the same procedure.
We use the Adam optimizer \cite{Kingma2014-xz} with a learning rate
of $3.10^{-4}$ and default momentum parameters. Networks are trained using
Pytorch and synchronous distributed training on $32$ gpus, with $32$ images per
GPU for a total batch size of $1024$. The networks are randomly initialized and we use the
following training procedure. First, the network is trained for $80$k
iterations on synthetic data only. Then, the network is trained for another
$80$k iterations on both real and synthetic training images. In this second
phase, the real training images account for around $25\%$ of each batch.
Following~\cite{Goyal2017-of},
we also use a warm-up phase where we progressively increase the learning rate from $0$
to $3.10^{-4}$ during the first $5$k iterations.

\subsubsection{Experimental findings.}
On YCB-Video, we found that pre-training the model on synthetic data yields an
improvement of approximately $2$ points on the AUC of ADD(-S) metric. Without this
pre-training phase, our model performed comparably to the results reported by
DeepIM. Note that this is hard to directly compare because the synthetic
training images are different from the ones used by DeepIM.

On T-LESS, we found that the data augmentation is crucial as also
pointed out by \cite{Sundermeyer2018-es}. Without data augmentation, the
performance of the coarse and refinement networks is poor, with a
$e_{vsd} < 0.3$ score of around 37\% compared to 64\% when training with data
augmentation.

\section{Object candidate matching: additional illustration}
\label{sec:supmat-matching}

\begin{figure}
  \centering
  \includegraphics[width=0.3\textwidth]{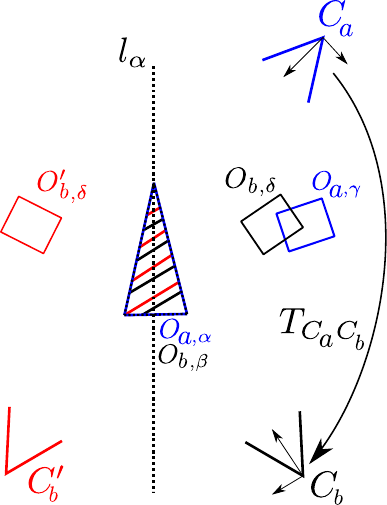}
  \caption{\small {\bf Relative camera pose estimation.} Given
  two pairs of object candidates $(O_{a,\alpha}, O_{b,\beta})$ and
  $(O_{a,\gamma}, O_{b,\delta})$, we estimate the relative camera pose
  $T_{C_aC_b}$ that best aligns candidates $O_{a,\gamma}$, $O_{b,\delta}$. In this
  example, the red camera pose $C_b^\prime$ is also valid due to the symmetries of the
  triangular object $l_{\alpha}$. It is discarded because the error between $O_{b,\delta}^\prime$ and $O_{a,\gamma}$ is bigger than between $O_{b,\delta}$ and $O_{a,\gamma}$.
}
  \label{fig:ransac}
\end{figure}

In Fig.~\ref{fig:ransac}, we illustrate our method for ``Sampling of relative camera poses
sampling'' described in Sec.~\ref{sec:method_matching} of the main paper with a simple 2D example.

\section{Scene refinement}
\label{sec:optim}

\subsubsection{Initialization.} There are multiple ways to initialize the
optimization problem defined in equation~\eqref{eq:ba_problem} of the main paper. We use the
following procedure. We start by picking a random camera and
setting it's coordinate frame as the world coordinate frame.
Then, we iterate over all cameras, trying to initialize each one. In order to initiliaze a
camera $a$, we randomly sample another camera $b$ which is already initialized (placed in the
world coordinate frame) and use the relative pose between these two cameras
$T_{C_aC_b}$ estimated while running RANSAC (relative camera pose sampling in
Sec. 3.2) to place camera $a$ in the world coordinate frame. Once all the
cameras have been initialized, we initalize objects by randomly picking an
object $p$ an initializing it using a candidate associated with this physical object from a random view.

\subsubsection{Rotation parametrization.} We use the same rotation parametrization
as the one used for our single-view single-object network for which the
equations are provided in Sec.~\ref{sec:singleview} of this appendix.

\section{Datasets and metrics}
\label{sec:datasetmetric}

\subsection{Datasets}

In this section, we give details of the datasets used in our experiments. 

\subsubsection{YCB-Video.} The YCB-Video \cite{Xiang2018-dv} dataset is made of
92 scenes with around 1000 images per scene.
The dataset is split into 80 scenes for training and 12 scenes for testing. It is mostly challenging due to the variations in lightning conditions, significant
image noise and occlusions. The objects are picked from a subset of 21 objects
from the YCB object set \cite{Calli2015-sz} for which reconstructed 3D models
are available. The models are presented in Fig.~\ref{fig:objects_ycbvideo}.
These models are used to generate additional synthetic training images.

\begin{figure}
  \centering
  \includegraphics[width=1.0\columnwidth]{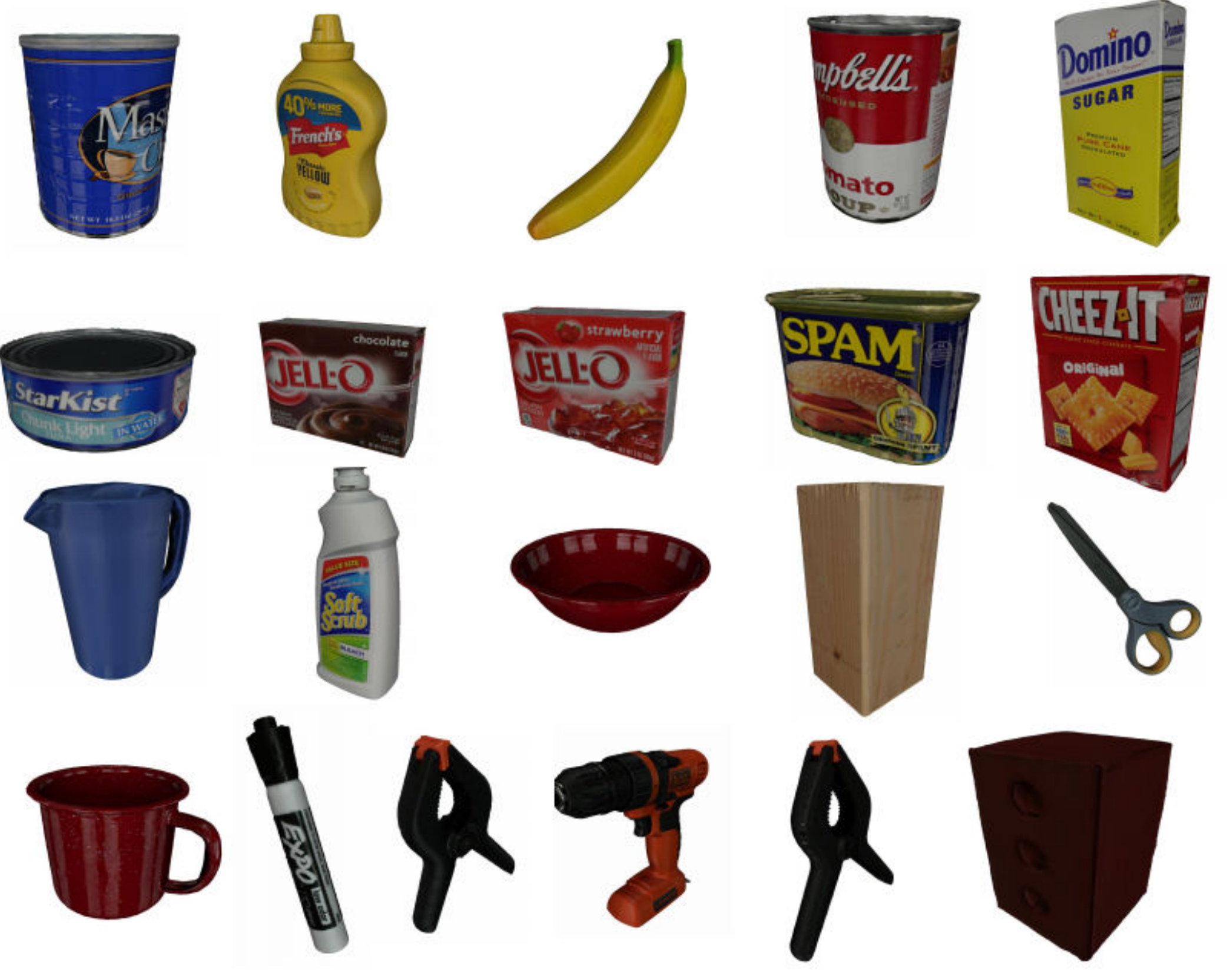}
  \caption{\small {\bf Objects of the YCB-Video dataset.} The 21 reconstructed
    object models of the YCB-Video dataset. Taken from \cite{Xiang2018-dv}.}
  \label{fig:objects_ycbvideo}
\end{figure}

There is at most one object of each instance per scene and most of the objects
are visually distinct with the exception of the large and extra-large clamps.
When testing, we follow previous works \cite{Li2018-fp,Xiang2018-dv,Peng2018-ev} and
evaluate on a subset of 2949 keyframes. The variety of the viewpoints for
each scene is limited as the camera is usually moved in front of the scene, but not completely around it.

\subsubsection{T-LESS.} The T-LESS \cite{Hodan2017-pq} dataset is made of 20 scenes featuring multiple
industry-relevant objects. There are 30 object instances, all of them are
textureless and most of them are symmetric.
The reconstructed 3D models of these objects are presented in Fig.~\ref{fig:objects_ycbvideo}.
Many objects have similar visual appearance, making the class prediction task challenging for the
object detector. The images in the dataset are taken all around the scene.
Scene complexity varies from 3 objects of different types to up 18 objects with 7 belonging to the same type. In
single-view experiments we consider all images of the testing scenes to provide meaningful
comparison with \cite{Sundermeyer2018-es,Park2019-od}. For multiview experiments
we consider the subset of the BOP19 challenge \cite{Hodan_undated-sl}. We use
the CAD models for generating synthetic images and for evaluation.

\begin{figure}
  \centering
  \includegraphics[width=1.0\columnwidth]{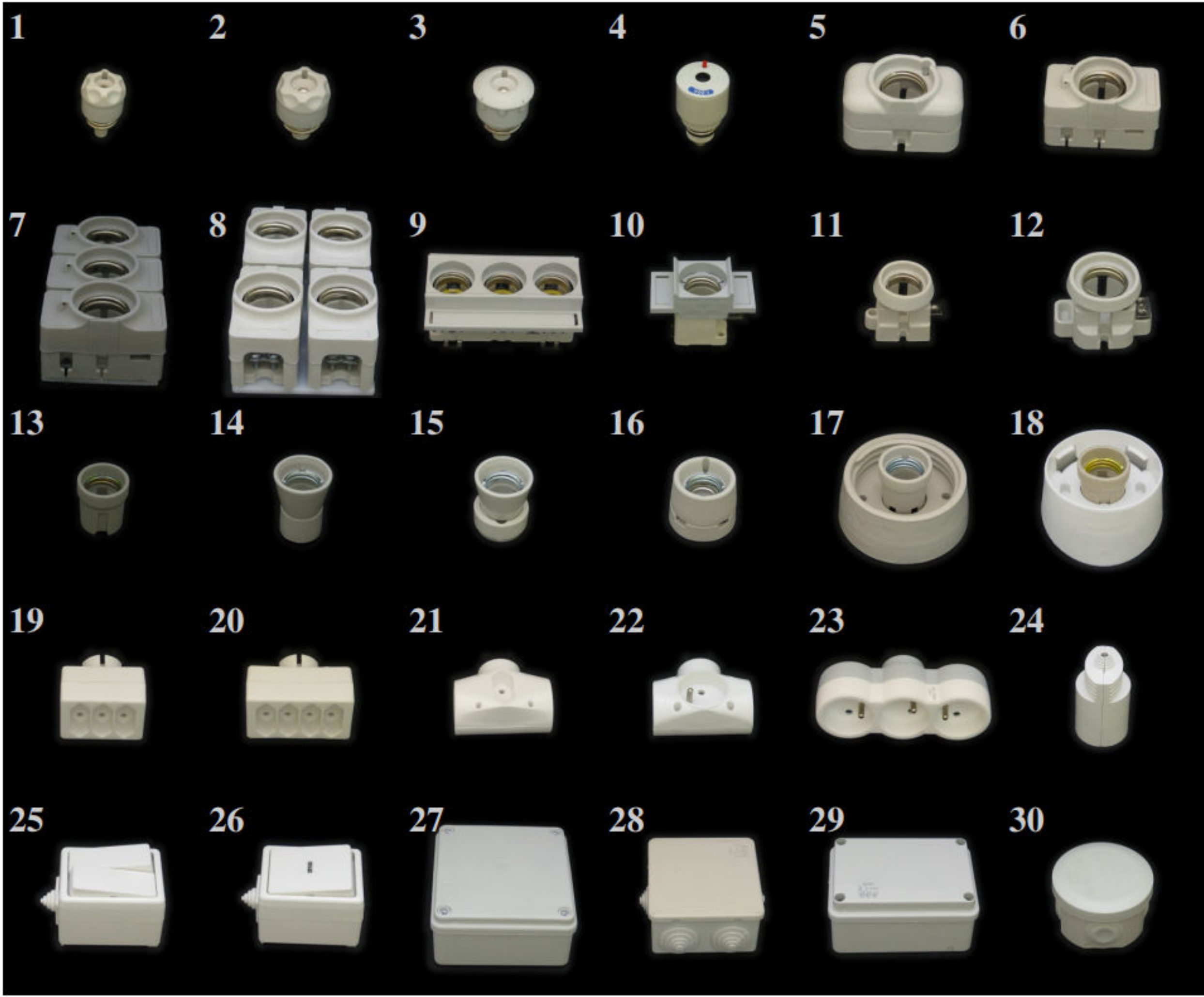}
  \caption{\small {\bf Objects of the T-LESS dataset.} The 30 reconstructed
    object models of the T-LESS dataset. Notice how multiple objects
    share visual appearances such as (1) (2); (5) (6); (14) (15) (16); (25) (26). Taken from \cite{Hodan2017-pq}.}
  \label{fig:objects_ycbvideo}
\end{figure}

\subsection{Metrics}
In this section, we give some details about the metrics reported in the main paper. We refer
to \cite{Hodan_undated-sl,Hinterstoisser2013-lx} for more information about
these metrics.

The ADD (average distance) metric is introduced in \cite{Hinterstoisser2013-lx}
and is typically used to measure the accuracy of pose estimation for non-symmetric
objects. Given a label $l$ of an object and following the notation introduced
in Sec.~\ref{sec:method_notations} of the main paper, this metric is computed as :

\begin{equation}
  \text{ADD}(l, T, \hat{T}) = \frac{1}{H_l} \sum_h ||\hat{T} X_l^h - T X_l^h||_2,
\end{equation}
where $T$ is the predicted object pose, $\hat{T}$ is the ground truth pose,
$X_l^h$ are the vertices of the 3D models and $H_l$ is the number of vertices of
the model of the object $l$.

For symmetric objects, the average distance is computed using the closest point
distance and noted ADD-S:
\begin{equation}
  \text{ADD-S}(l, T, \hat{T}) = \frac{1}{H_l} \sum_h \min_{g} ||\hat{T} X_l^h - T X_l^g||_2.
\end{equation}

The notation ADD(-S) corresponds to computing ADD for non symmetric objects and
ADD-S for symmetric objects. It is also common to report the percentage of objects
for which the pose is estimated within a given threshold such as 10\% of it's
diameter. We use the notations ADD-S $<$ 0.1d and ADD(-S) $<$ 0.1d for this
metric and report the mean computed over object types.

The authors of PoseCNN \cite{Xiang2018-dv} also proposed to report the area under the
accurracy-threshold curve for a threshold (on ADD-S, or ADD(-S)) varying between
0 to 10cm. We note this metric as AUC of ADD(-S) or AUC of ADD-S and we use the
implementation provided with the evaluation code\footnote{https://github.com/yuxng/YCB\_Video\_toolbox} of YCB-Video.

{When evaluating on the T-LESS dataset, we also report the Visual Surface Discrepancy metric
(vsd). This metric is invariant to object symmetries and takes into account
the visibility of the object. As in \cite{Sundermeyer2018-es,Park2019-od}, the
pose is considered correct when the error is less than 0.3 with $\tau=20mm$ and
$\delta=15mm$. We note this metric $e_{\text{vsd}}<0.3$ and use the official
implementation code of the BOP challenge
\cite{Hodan_undated-sl}\footnote{https://github.com/thodan/bop\_toolkit}. There
are multiple instances of objects in multiple scenes of the T-LESS dataset.
When comparing with prior work \cite{Sundermeyer2018-es,Park2019-od} on all
images of the primesense camera, we only evaluate the prediction which has the
highest detection score for each class, and only objects visible more than 10\%
are considered as ground truth targets. This corresponds to the SiSo task.

When evaluating our multi-view method, we follow the more recent 6D localization
protocol of the ViVo BOP challenge which considers the top-$k$
predictions with highest score for each class in each image, where $k$ is the
number of ground truth objects of the class in the scene. Note that the metrics of the BOP
challenge do not penalize making many incorrect predictions for classes
that are not in the scene, which happens in most methods and is problematic
for practical application. We thus propose to analyze precision-recall
tradeoff similar to the standard practice in object detection, using ADD-S$<$0.1d
to count true positives.

When computing the mean of ADD-S errors in our scene refinement ablation,
we only consider as true positives predictions the ones which have an ADD-S error lower than half of the diameter of the
object, to ensure that the prediction is matched to the correct ground truth
object. Without limiting the error to this threshold and using only class labels
and scores, some predictions may be matched to ground truth objects which are
at a very different location in the scene. This tends to increase the errors while not being
representative only of the 6D pose accuracy of the predictions.
}

\section{Additional multi-view multi-object results}
\label{sec:qualitative_examples}
Each scene reconstruction is presented with a dedicated figure and we provide close-ups on various parts of the visualization to illustrate the different aspects in detail. The explanation is
provided in the caption of each figure. 

\subsubsection{Layout of the figures.} In each figure presented below, four (on
T-LESS) or five (on YCB-Video) RGB images were used to reconstruct each scene.
In each figure, each row corresponds to results associated with one image and different columns present the results of different stages of
our method. The last column shows the ground truth scene. The different columns are described next.
\begin{itemize}
\item ``Input image'' is the (RGB) image used as input to the method.
\item ``2D detections'' shows the detections obtained by the object
  detector (RetinaNet on T-LESS, PoseCNN on
YCB-Video), after removing detections that have scores below $0.3$. The
color of each 2D bounding box illustrates the object label predicted for this
detection, each color is associated with a unique type of 3D object in the object database.
Note that the colors for each type of 3D object are shared for all visualizations
corresponding to one scene (one figure) but not shared across the figures
because of the high number of objects in the database.
\item ``Object candidates'' illustrates the 6D object poses predicted for each
 2D detection. The candidates considered as outliers (those who have not been
 matched with a candidate from another view and are discarded) are marked with
 red color and are transparent.  The candidates considered inliers are shown in green. Inliers are used in the final scene reconstruction. 
 Note that the red and green colors in this (3rd)
 column are only used to indicate inliers and outliers and there is no
 correspondence with red and green colors in the 4th column that denote the different object types.
\item ``Scene reconstruction'' illustrates the scene reconstructed by our method
  using all the views presented in the figure. Once the scene is reconstructed,
  we use the recovered 6D poses of physical objects and cameras to render
  the scene imaged from each of the predicted viewpoints. The
  renderings are overlaid over the input image.
\item ``Ground truth'' corresponds to the ground truth scene viewed from the
 ground truth viewpoints. These images are shown to enable visual comparison with the results of our method. {The ground truth information (number of objects,
  types of objects, poses of cameras, poses of objects) is not used by our method.}
\end{itemize}

In the following, we illustrate the main capabilities of our system. 

\subsection{Highlights of the capabilities of our system}

\subsubsection{Large number of objects, robustness to occlusions, symmetric objects.} 
\begin{figure}
  \centering
  \includegraphics[width=1.0\columnwidth]{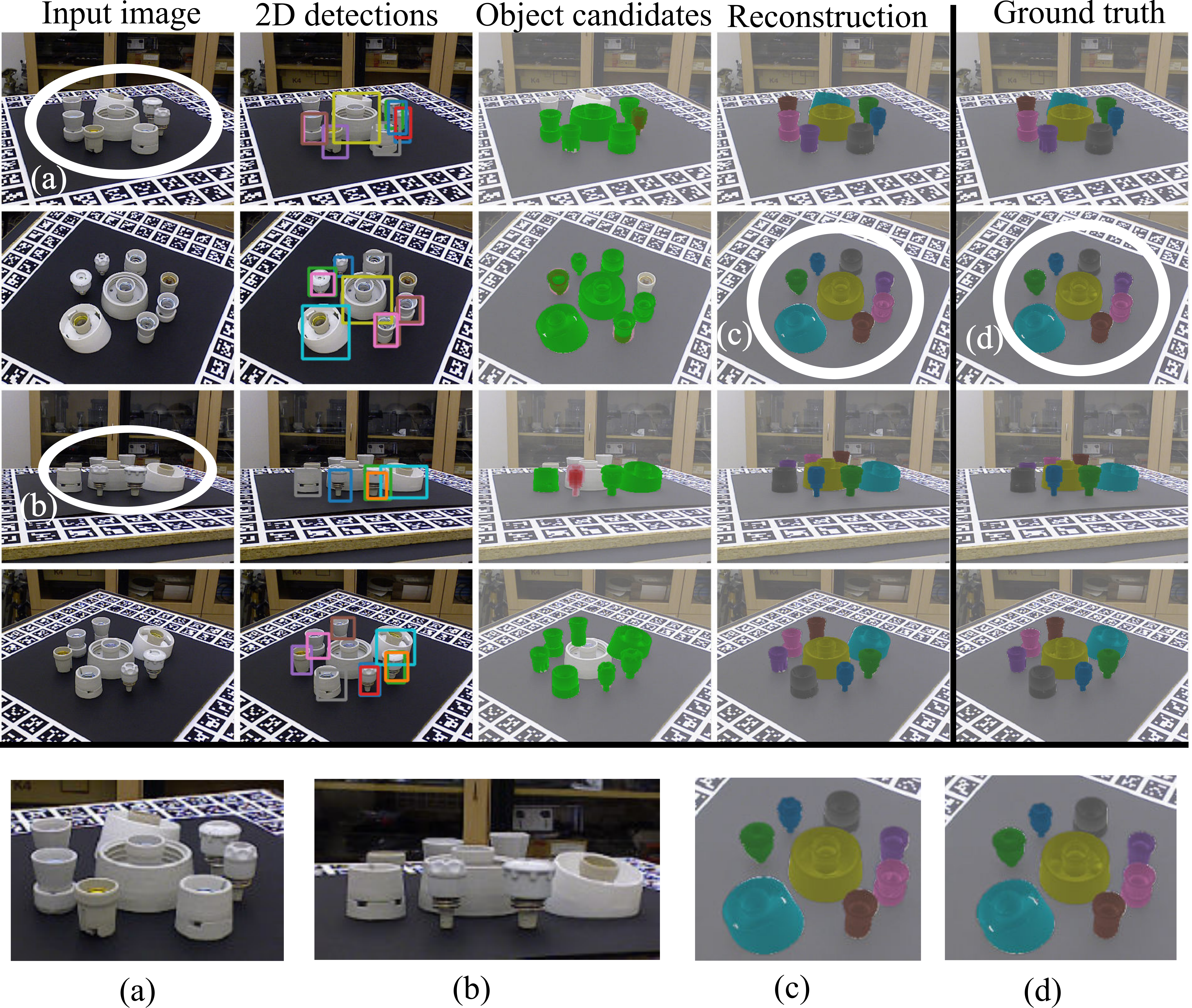}
  \caption{\small {\bf Highlight I: Scene with many symmetric objects and occlusions.} Our method is able
    to correctly identify and predict the poses of the 8 symmetric objects present in the scene. Please note how object poses
    and labels/colors are similar in the output of our method, shown in close-up (c), and the ground
    truth, shown in close-up (d). This is particularly challenging because of the high object
    density, varying level of occlusions and the fact that all objects of the
    scene are symmetric, as shown in close-ups (a) and (b).}
  \label{fig:many_objects}
\end{figure}

Our method is
able to recover the state of complex scenes that contain multiple objects, even
if parts or the scene are partially or completely occluded in some of the views.
The poses of cameras and objects can be correctly recovered even if all objects
in the scene are symmetric. An example is presented in Fig.~\ref{fig:many_objects}. Note how some objects are missing in each individual view but our method is able to recover correctly all objects.

\subsubsection{Multiple object instances.} 
\begin{figure}
  \centering
  \includegraphics[width=1.0\columnwidth]{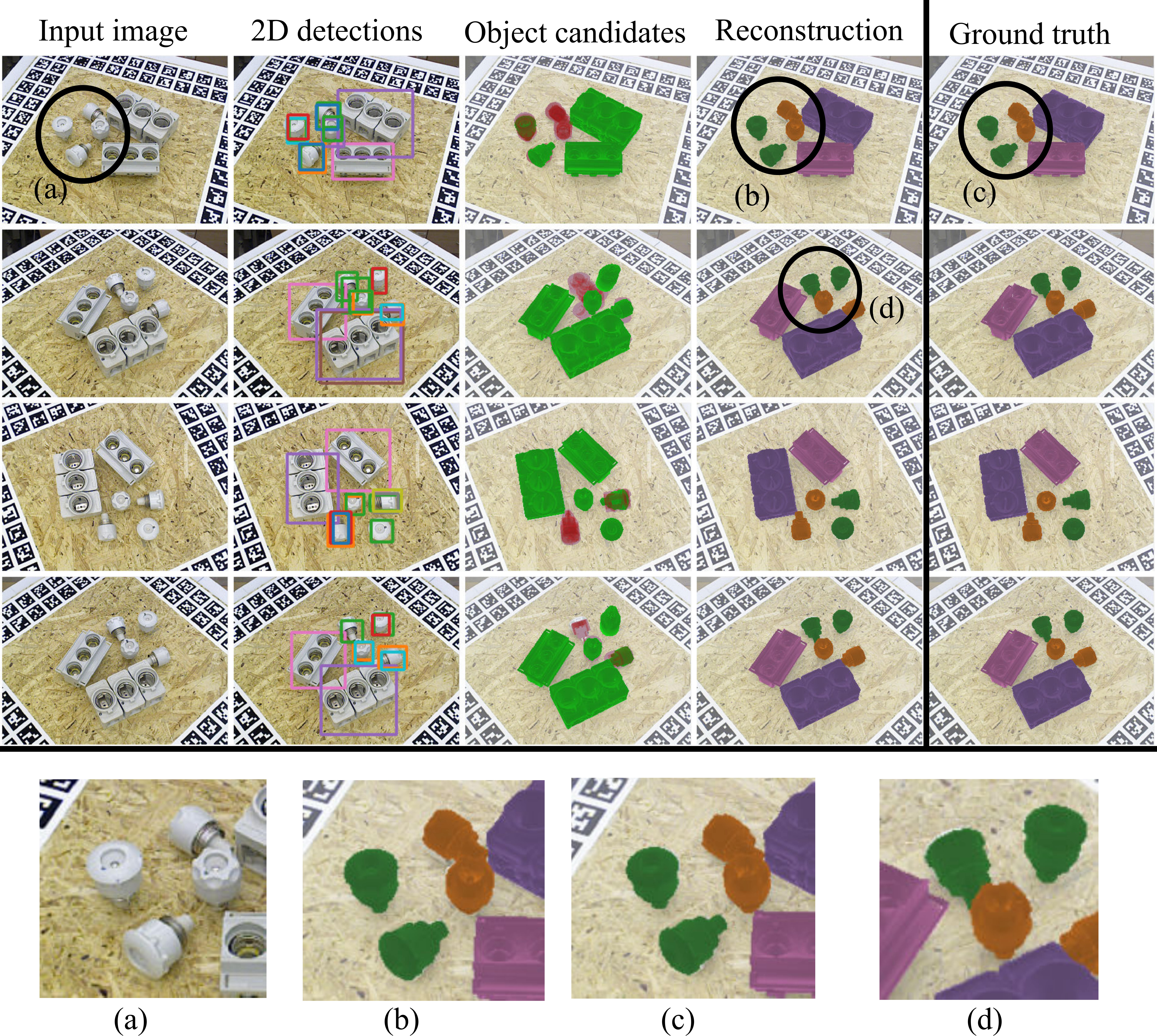}
  \caption{\small {\bf Higlight II: Scene with multiple object instances of the same object type.} Note how our 
    method is able to correctly identify all objects in this challenging scene. Object poses
    and labels/colors predicted by our method, shown in close-up (b) are very similar to the ground truth, shown in close-up (c). This is particularly challenging because the green and orange
    objects have similar visual appearance, are close to each other in the
    scene, and objects are partially occluded in some of the views, as shown in close-ups (a) and (d).
  }
  \label{fig:multi_instance}
\end{figure}

Our method is able to
successfully identify the correct number of objects and their labels
even if there are multiple objects of the same type in the image, objects
are partially occluded in some views and multiple types of objects have very
similar visual appearance. An example is presented in Fig.~\ref{fig:multi_instance}

\subsubsection{Cluttered scenes with distractors.} 
\begin{figure}
  \centering
  \includegraphics[width=1.0\columnwidth]{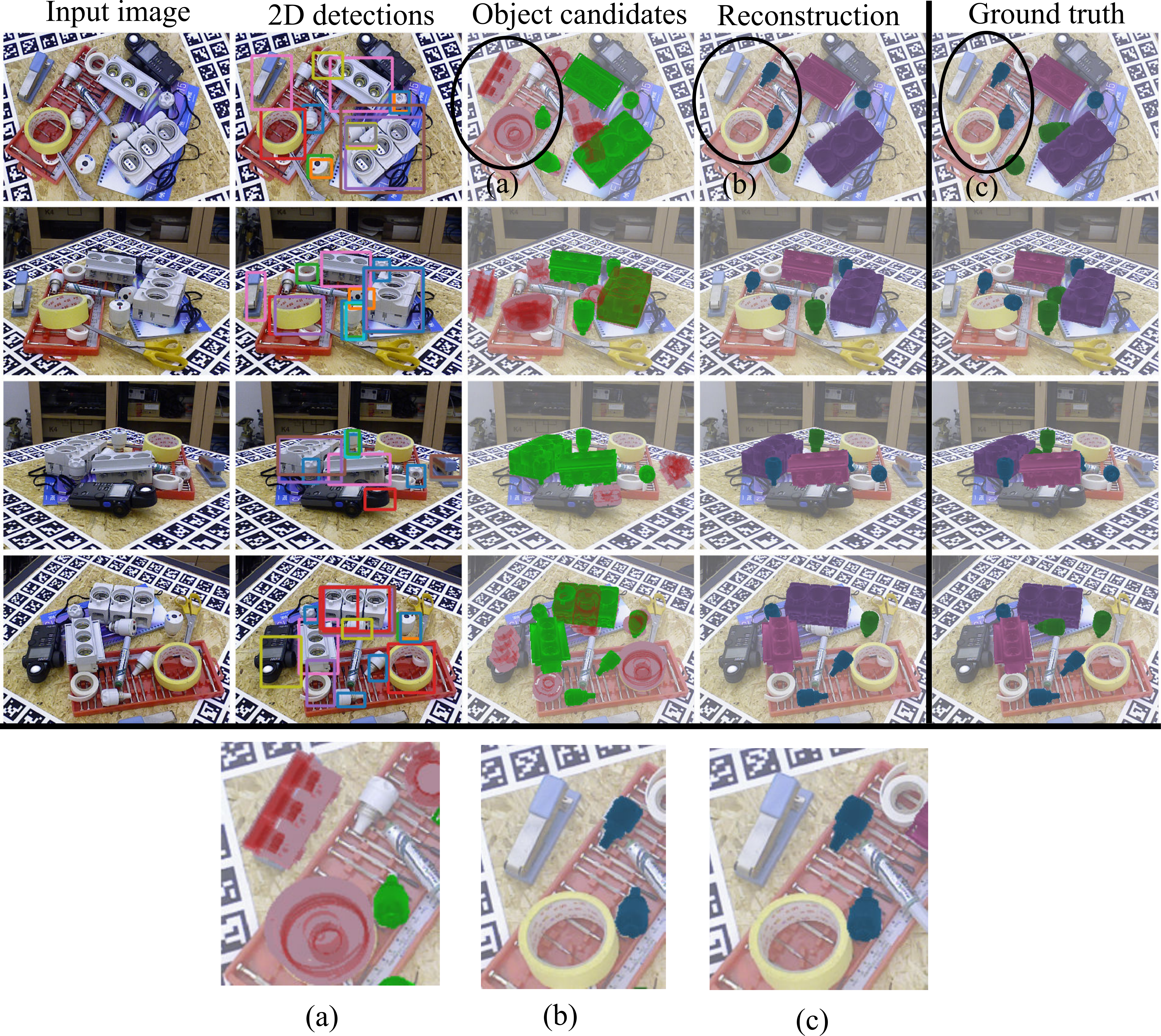}
  \caption{\small {\bf Highlight III: Scene with multiple distractors.}
  Our method is also robust to
distractor objects that are not in the database of objects. Our method correctly localizes and estimates the pose of all databse objects in the scene (cf. our reconstruction (4th column) and the ground truth (5th column)) despite the presence of several distractor objects (objects not colored in the ground truth).  A single-view approach (Object candidates, 3rd column) incorrectly detects three of the distractor
    objects and places them in the scene because they look similar to some
    objects of the database, as shown in the close-up (a). 
    Our robust multi-view approach is able to
    filter these outliers: the objects estimated at the positions of the
    distractors are marked in red in (a). Distractor objects have been filtered
    in the final reconstruction as shown in the close-up (b) (cf. ground truth close-up (c)).}
  \label{fig:distractors}
\end{figure}

Our method is also robust to
distractor objects that are not in the database of objects. We present in
Fig.~\ref{fig:distractors} a complex example with many distractors where
our method is able to successfully recover all objects in the scene, which are
in the object database while filtering out the other ones. This is especially
important for robotic applications in unstructured environments where the
objects of interests are known and should not be confused with other background objects.

\subsubsection{High accuracy.} 
\begin{figure}
  \centering
  \includegraphics[width=1.0\columnwidth]{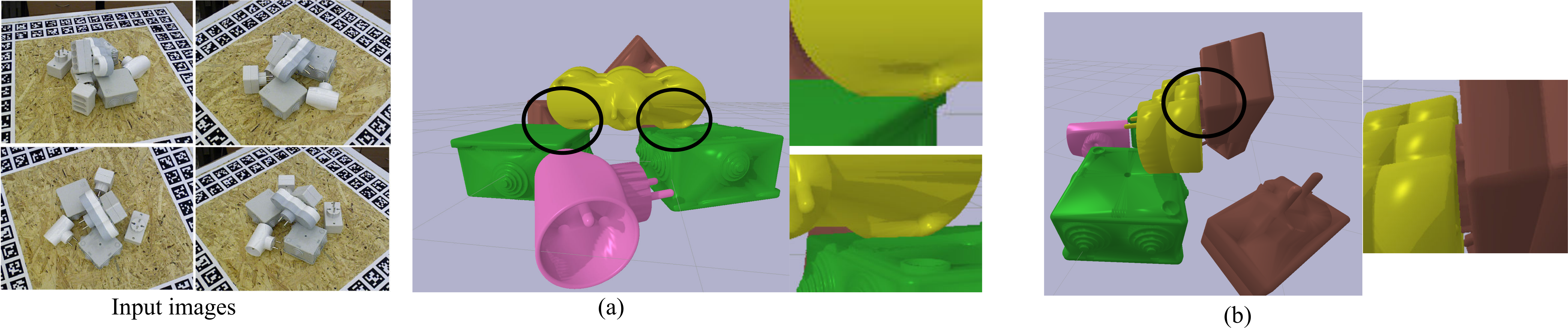}
  \caption{\small {\bf Highlight IV: Accuracy of our approach.} Left: input images. Then (a) and (b) shows the output scene imaged from two viewpoints different from the views used for the reconstruction. Please note in (a) how  the yellow object is accurately estimated to only touch the green objects, and in (b) how the brown object is correctly plugged inside the yellow object.}
  \label{fig:high_accuracy}
\end{figure}

One of the key components of our approach is
scene refinement (section~\ref{sec:method_ba} in the main paper), which significantly improves  the accuracy of pose predictions using information from multiple views. In Fig~\ref{fig:high_accuracy}, we show an
example of a reconstruction that highlights the accuracy that can be reached by
our method using only 4 input images.

\subsection{Detailed examples}

We now explain in detail few simpler examples that demonstrate how our system works and how it achieves the kind of results presented in the previous section.

\subsubsection{Robustness to missing detections.} 
\begin{figure}
  \centering
  \includegraphics[width=1.0\columnwidth]{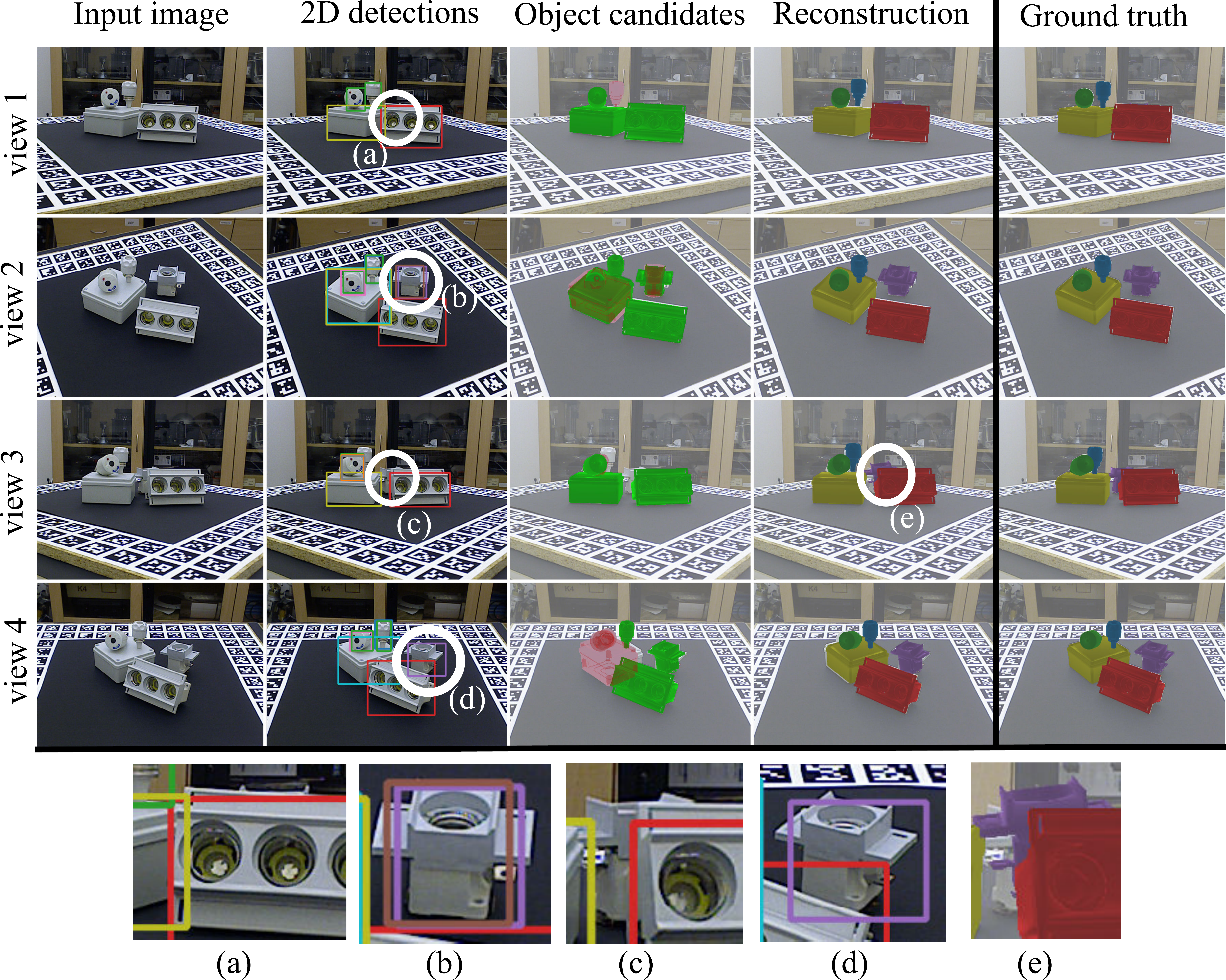}
  \caption{\small {\bf Example I: robustness to missing detections.} One of the
    objects (marked by purple circle) in the scene is detected in two views (b) (d), but not in the other two views due to
    partial (c) or complete (a) occlusion. Our method is able to (i) position the
    views 1 and 3 with respect to the scene using the other visible candidate objects
     and (ii) position the purple object with respect to these other
    objects using views 2 and 4, where the purple object is visible. Once the scene is reconstructed, it is also
    possible to directly recover the pose of the purple object with respect to
    views, where it was not originally detected, like in (e). 
    }
  \label{fig:missing_detections}
\end{figure}

In some situations, objects are
partially or completely occluded in some of the views. As a result, 2D
detections for one physical object are missing in some views.
If this physical object is visible in other views, our reconstruction method
is able to estimate it's pose with respect to the other objects. If all
cameras can be positioned with respect to the rest of the scene using other
non-occluded objects, our approach can also position the partially occluded
object with respect to all cameras, even if there were initially no candidates
corresponding to the object in these views. An example is shown in Fig.~\ref{fig:missing_detections}.

\subsubsection{Robustness to incorrect detections.} 
\begin{figure}
  \centering
  \includegraphics[width=1.0\columnwidth]{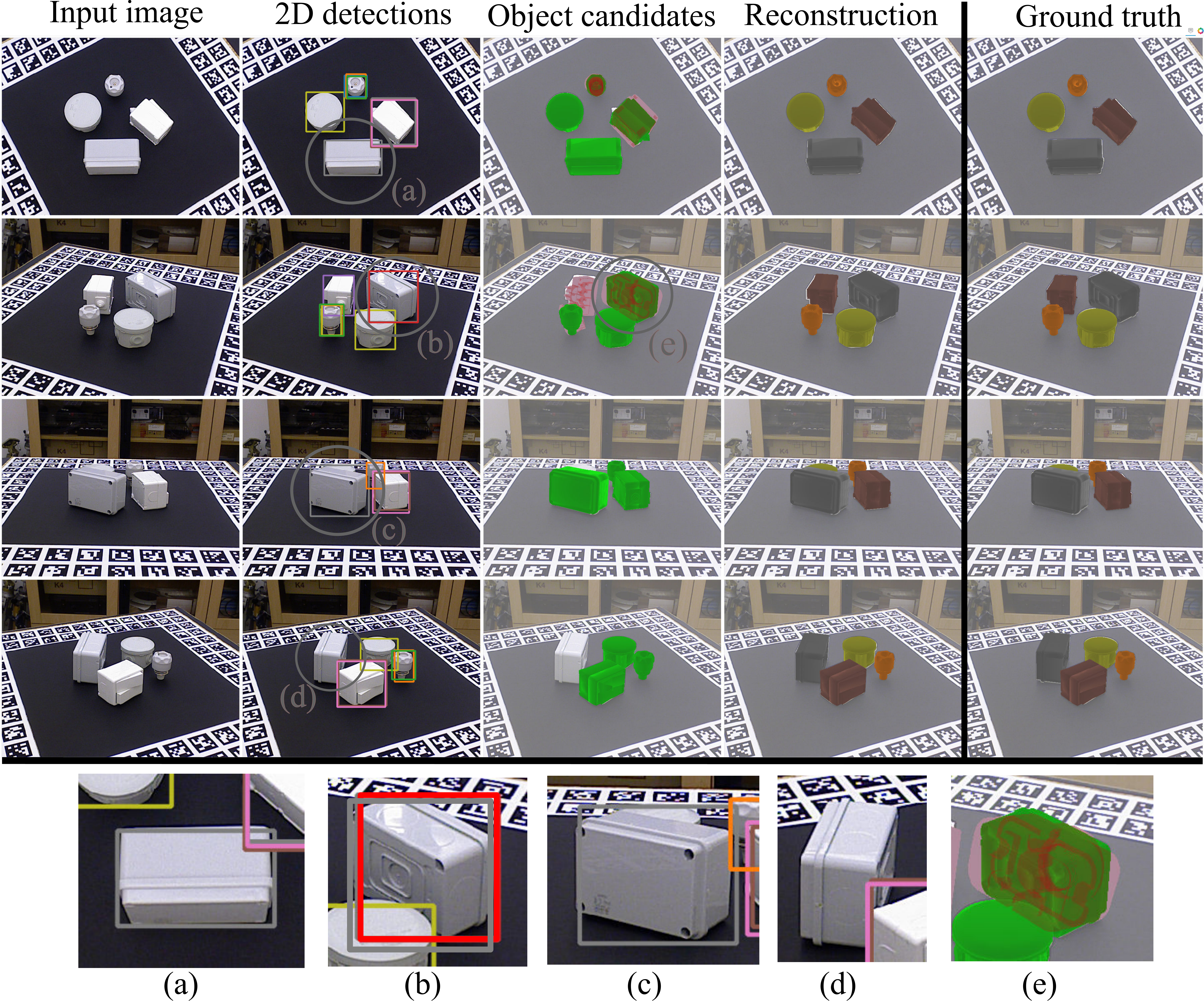}
  \caption{\small {\bf Example II: Robustness to incorrect detection labels.} One of the
    objects that is correctly identified in two views (a) (c), has two label
    hypotheses in view (b) and is not detected in view (d). Our
    method keeps the two hypotheses in (b) and predicts two 6D object candidates
    (e) but it is able to discard one of them because it's label is not consistent with
    the other views: one of the two object candidates is marked as an outlier (red) in
    (e). In our final scene reconstruction, the gray object is
    correctly recognized (it has the same color (gray) in out output ``Reconstruction'' and in
    the ``Ground truth'').}
  \label{fig:wrong_labels}
\end{figure}

In T-LESS, many
objects have similar visual appearance. As a result, the 2D detector often makes
mistakes, predicting incorrect labels for some of the detections in some views.
Our method is able to handle multiple 2D detections that have different labels
at the same location in the image. In this case, a pose hypothesis is generated for
each of the label hypothesis. If the object candidate cannot be matched with
another view - either because the incorrect label is predicted in only one view
or because the poses are not consistent - our method is able to discard this
object candidate. An example is shown in Fig.~\ref{fig:wrong_labels}.
Please see the discussion ``Duplicate objects'' and Fig.~\ref{fig:duplicates}
for examples where an object is consistently mis-identified across multiple views.

\subsubsection{Duplicate objects.} 
\begin{figure}
  \centering
  \includegraphics[width=1.0\columnwidth]{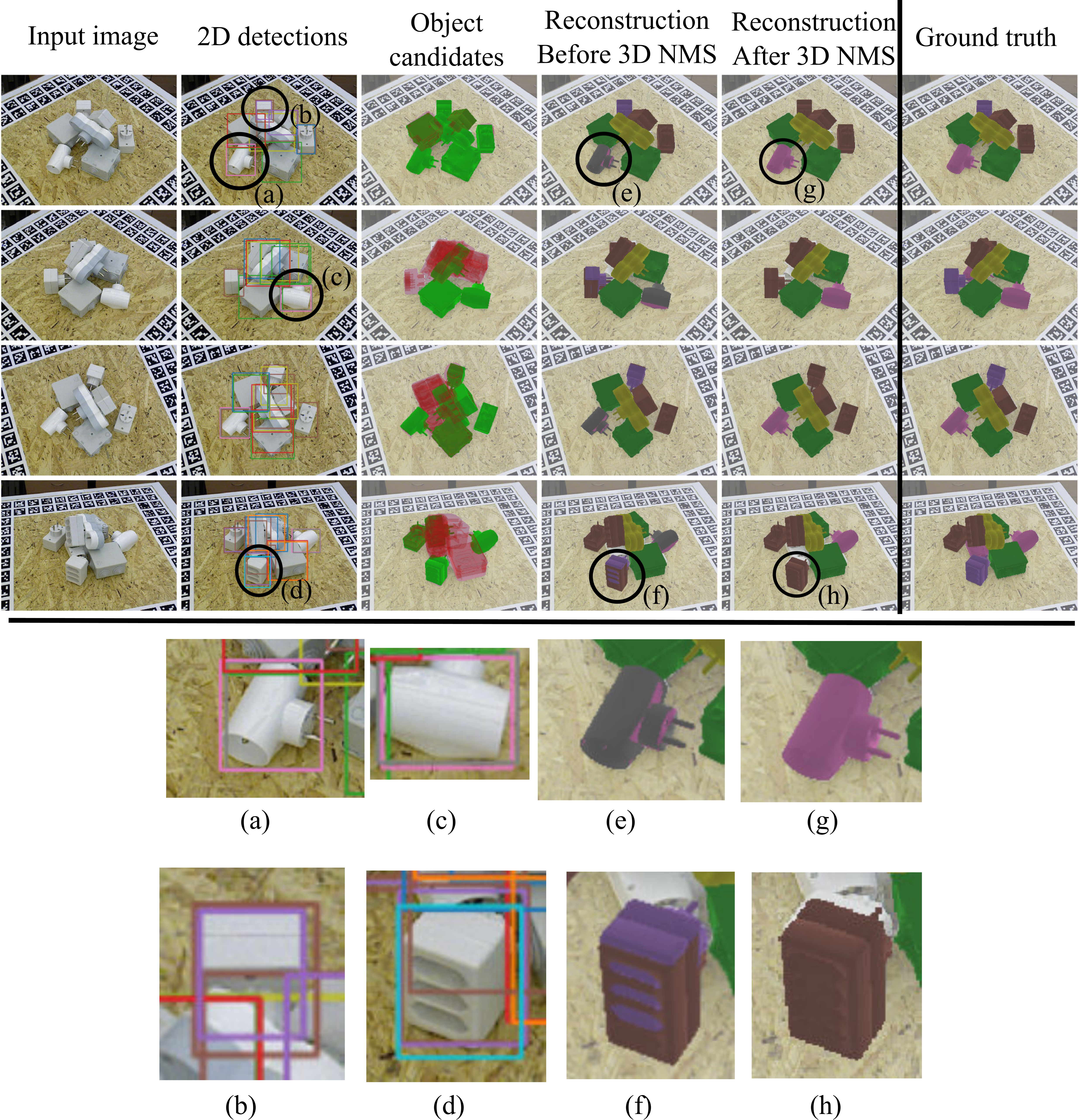}
  \caption{\small {\bf Example III: Duplicate objects.} 2D detections with two different
    labels (grey and pink) are predicted for the same object consistently across
    two views, (a) and (c). Because the 3D models of the pink and grey objects are similar, the poses predicted in both views
    are consistent and thus both pairs of object candidates are associated
    to separate objects. In the final scene reconstruction, two objects (grey
    and pink) overlap at the same 3D location (e). We use a 3D
    non-maximum suppression strategy to retain only a single hypothesis. In the final output (after NMS), the
    correct object is retained (pink), c.f. the ground truth column.
    In some cases, incorrectly identified objects are kept as shown in (b), (d), (f), (h).
  }
  \label{fig:duplicates}
\end{figure}

When multiple objects share the same visual
appearance as it is the case in the T-LESS dataset, there are often multiple label
hypotheses that are consistent across views for the same physical object.
Because these objects look similar to each other and match the observed
image, the pose estimation network (which tries to match a rendering with the
observed image, regardless of the object type) predicts reasonable poses for each label
that are consistent across different views. These candidates are matched across views
and multiple objects with different labels are predicted in the final scene at
the same spatial position. In our visualization, we remove these
duplicate objects by using a simple 3D non-maximum suppression (NMS) strategy on the estimated
physical objects of the final scene. If multiple objects are too close to each
other in the 3D scene, we keep the object with the highest score --
 the sum of the 2D detection scores of all inlier object candidates that are
associated with one physical 3D object. Duplicate objects and 3D non-maximum
suppression are illustrated in Fig.~\ref{fig:duplicates}, including one correct and one incorrect example. The column ``Reconstruction'' in all figures corresponds to the output of our method
\textit{after} the 3D NMS.

\subsubsection{Robustness to distractors and false positives.} 
\begin{figure}
  \centering
  \includegraphics[width=1.0\columnwidth]{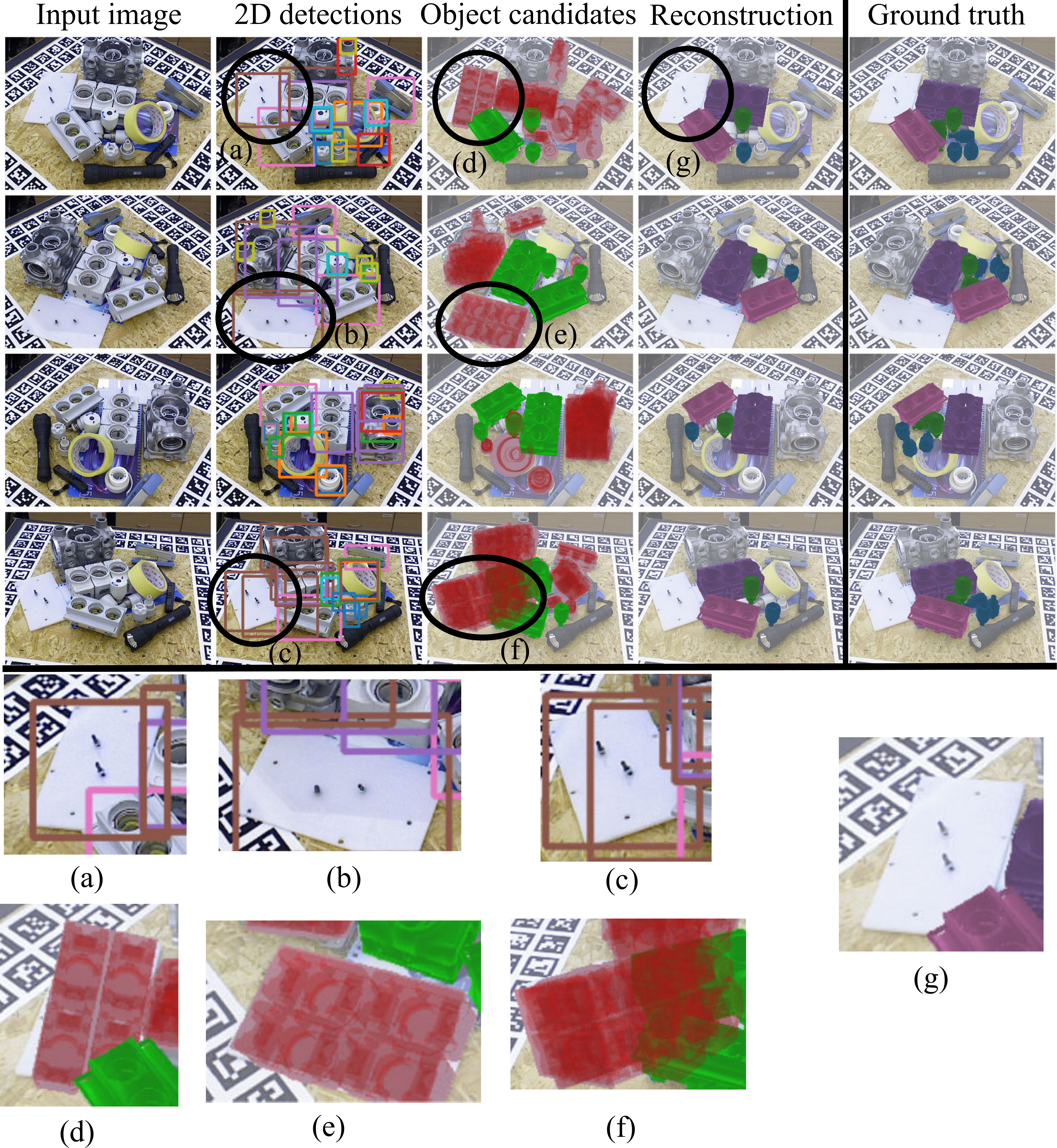}
  \caption{\small {\bf Example IV: Robustness to false positive detections.} One of the
    distractor objects is incorrectly detected in three views, see close-up (a), (b) and (c), with a consistent
    label (brown). For each of these detections, a 6D object candidate is
    generated, see close-ups (d), (e) and (f), but the poses are inconsistent across views because the pose
    estimation network has not been trained for this object. These candidates
    are filtered by our robust candidate matching strategy and considered
    outliers (red), see (d),
    (e) and (f). Note how this distractor is not present in the final scene
    reconstruction, as shown in close-up (g).}
  \label{fig:false_positive_success}
\end{figure}

The complex scenes in the T-LESS dataset
also have background distractor objects that are not in the object database.
Some of these distractors look similar to objects in the database and can be incorrectly detected, sometimes in multiple images. In these cases, the pose
estimator most often produces 6D pose estimates that are not consistent across views
because the input real images are outside of the training distribution (they
display objects that are not used to generate the training data). Because these
estimates are not consistent across views, our method is able to filter them and
mark them as outliers (red), thus gaining robustness with respect to these distractors. An
example is shown in Fig.~\ref{fig:false_positive_success}. 

\subsection{Limitations}

We now describe the most challenging scenarios that our method is currently not able to
recover from. For each of these, we briefly discuss possible improvements.

\subsubsection{Limitation I: consistent mistakes} 
\begin{figure}
  \centering
  \includegraphics[width=1.0\columnwidth]{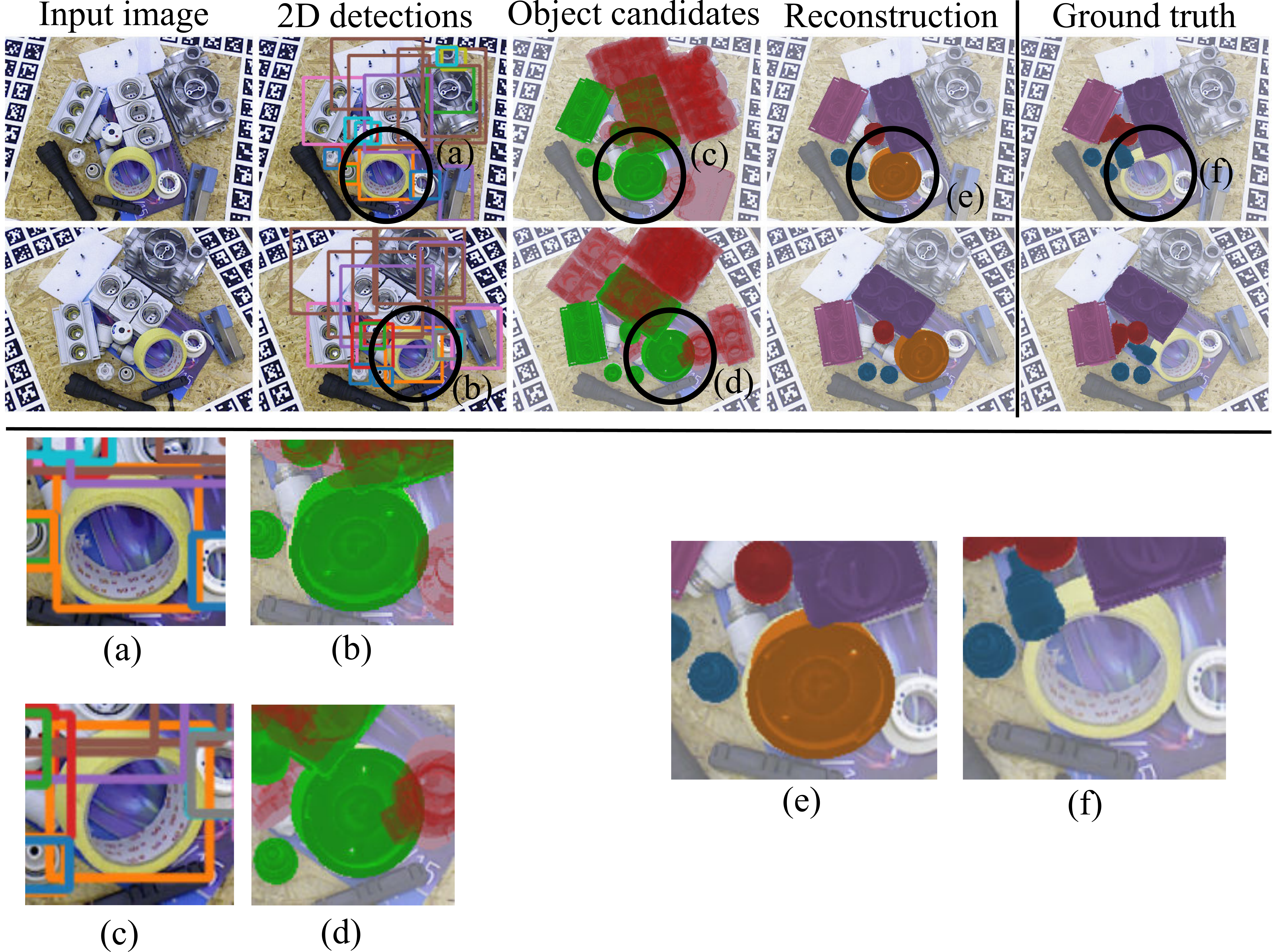}
  \caption{\small {\bf Limitation I: Consistent mistakes.} One of the
    distractors is incorrectly detected as an orange object (from the object
    database), as shown in close-ups (a) and (c). The two viewpoints are quite similar and as a result  the two estimated
    object poses are consistent, as shown in (c) and (d). The object is present in the final
    reconstruction (e) but it does not correspond to the ground truth object (f).}
  \label{fig:consistent_mistake}
\end{figure}

If two incorrect 6D object
candidates are consistent across at least two views, an (incorrect) object will be
present in the reconstructed scene. Such failure case typically happens when two
viewpoints are similar to each other. An example is shown in
Fig.~\ref{fig:consistent_mistake}. If two views are very similar, the
incorrect candidates will be matched together. Note that this failure mode could
be resolved by using a higher number of views, and by only considering physical
objects that have a sufficiently high number of associated object candidates. 

\subsubsection{Limitation II: Objects missing in the final reconstruction.} 
\begin{figure}
  \centering
  \includegraphics[width=1.0\columnwidth]{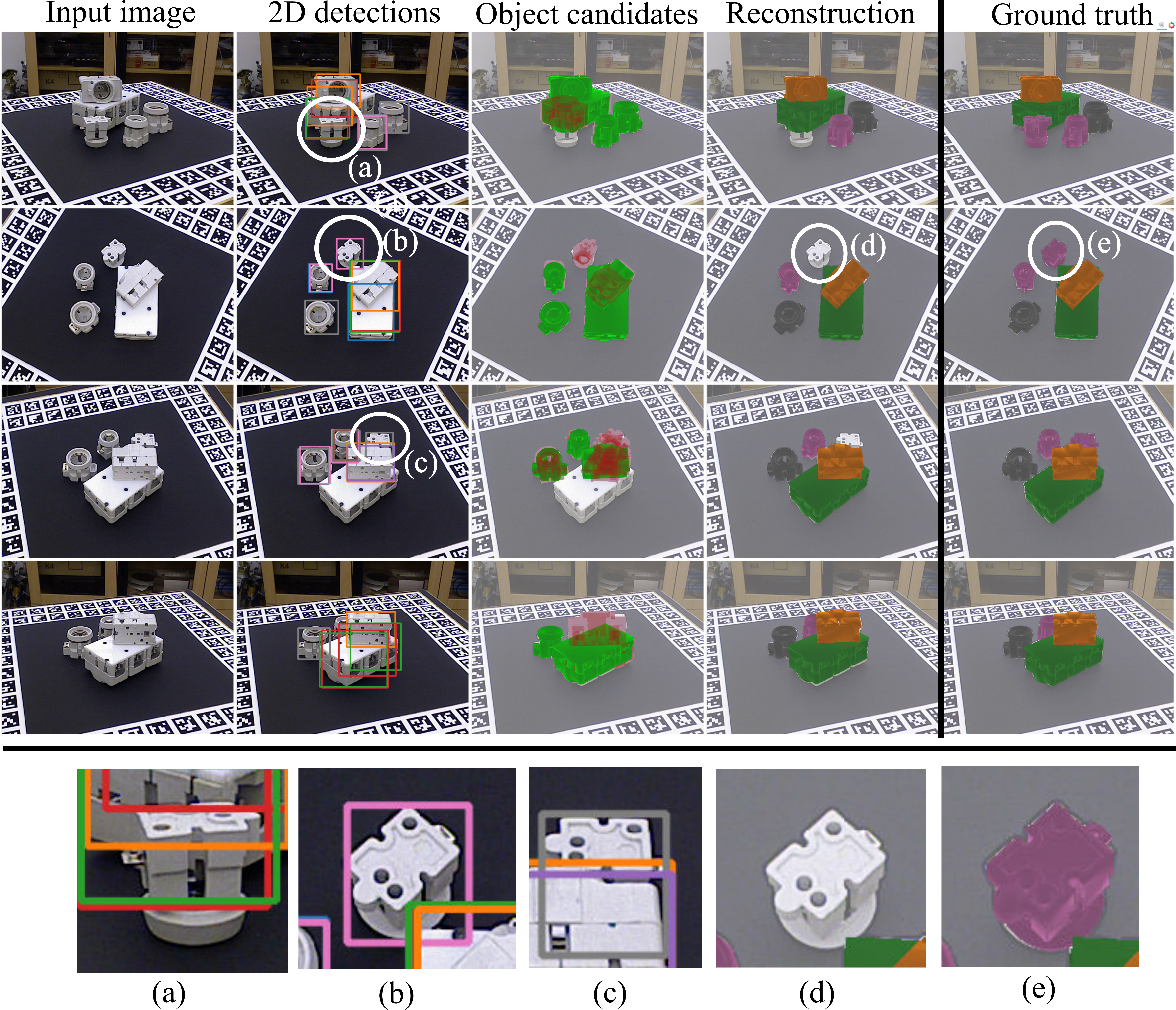}
  \caption{\small {\bf Limitation II: missing objects.} An
    object is detected correctly in one view as shown in the close-up (b), but the detection is missing in
    other views, shown in close-up (a), or the detection is incorrect and inconsistent, as shown in close-up (c). The
    object candidate (b) cannot be matched with another candidate and thus is
    missing from the final reconstruction, as shown in close-up (d) of the output (cf. ground truth close-up (e)).}
  \label{fig:missing_object}
\end{figure}

Our
current approach requires that a candidate in one view is matched with at least
one candidate from another view. If a candidate detection and pose estimate is
correct in one view but not in any other view, it will be missing from the final
reconstruction. An example is presented in Fig.~\ref{fig:missing_object}. Note
that in this case, all camera poses are still estimated correctly. An
interesting direction to overcome this problem  would be to grow the number
of object candidates in each view by reprojecting the detection from other views, as done in guided matching.

\subsubsection{Limitation III: Incorrect estimates of camera pose.} 
\begin{figure}
  \centering
  \includegraphics[width=1.0\columnwidth]{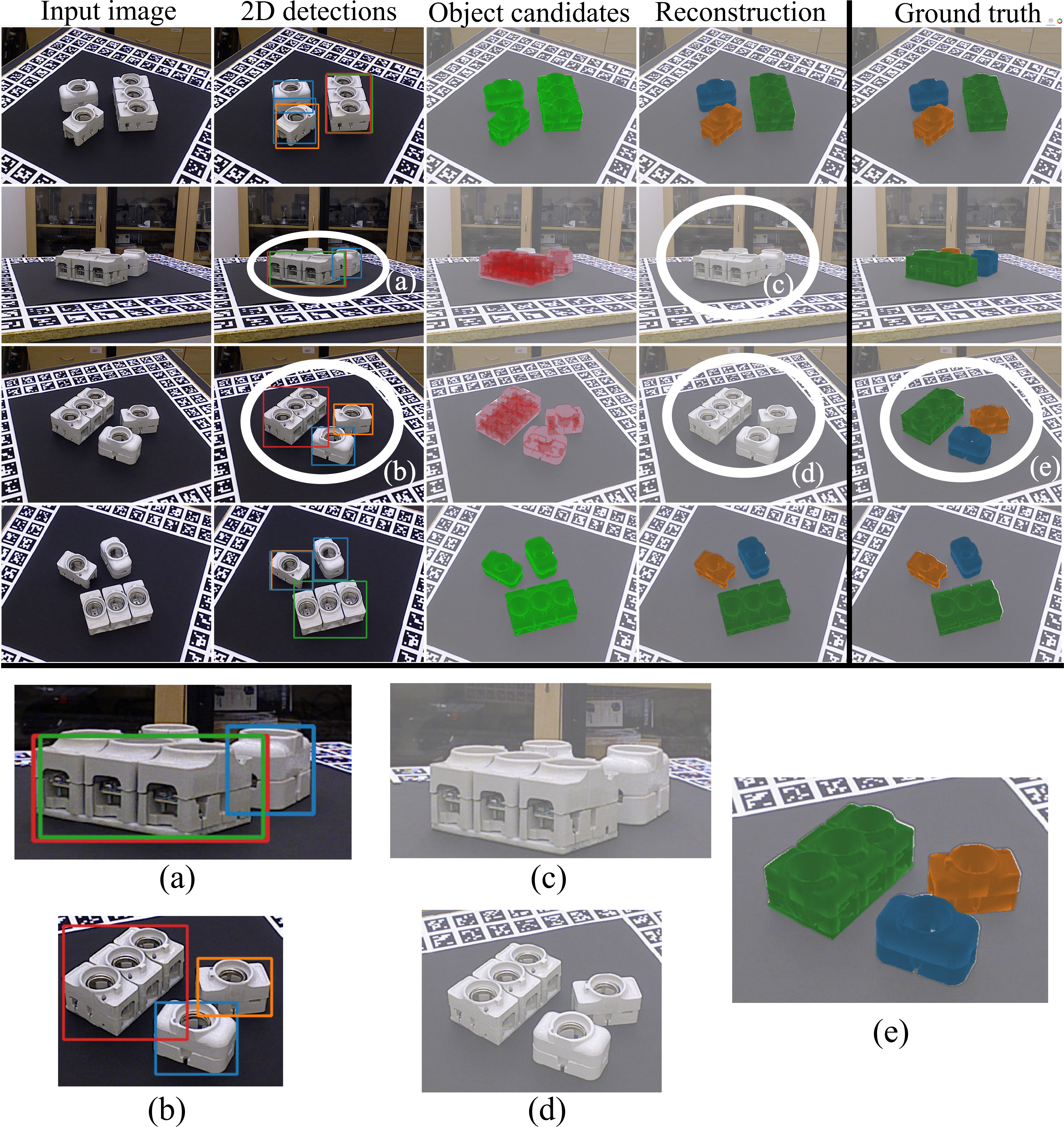}
  \caption{\small {\bf Limitation III: incorrect estimates of camera pose}. If one
    view has only two visible objects, as shown in close-up (a), the corresponding camera view with respect
    to the rest of the scene cannot be estimated as it requires at least three correctly estimated objects. As a result the objects are not reprojected in the
    image (c). This also happens if three candidates are detected in one view, as shown in close-up
    (b), but one of the object candidates is not consistent with the other views (here red object instead of green object).
}
  \label{fig:missing_camera}
\end{figure}

To position the
camera with respect to the scene, our method requires that there are at least
three object candidate inliers in the view: two for positioning the camera with
respect to the scene, and another one to validate the camera pose hypothesis.
Sometimes, however, there is insufficient number of inliers. This typically happens if only two objects are visible, or if there is a small
number of objects visible and some of the detections are incorrect. An example
is shown in Fig.~\ref{fig:missing_camera}.

\end{document}